\documentclass[10pt,twocolumn,letterpaper]{article}

\usepackage{cvpr}              % To produce the CAMERA-READY version
\definecolor{cvprblue}{rgb}{0.21,0.49,0.74}
\usepackage[pagebackref,breaklinks,colorlinks,allcolors=cvprblue]{hyperref}

\usepackage{booktabs} % For formal tables
\usepackage{multirow} % For multi-row table headers
\usepackage{amsmath}  % For align, equation environments used in method section
\usepackage{threeparttable}
\usepackage{float}
\usepackage{tikz} % For method pipeline diagram
\usetikzlibrary{arrows.meta,calc,decorations.pathreplacing,positioning}
\usepackage{subcaption}
\usepackage{graphicx}
\usepackage[table]{xcolor}
\newcommand{\best}[1]{\underline{#1}}
\definecolor{color_1st}{rgb}{0.82, 0.71, 0.58}
\definecolor{color_2nd}{rgb}{0.945, 0.87, 0.74}
\definecolor{color_3rd}{rgb}{1, 0.98, 0.84}
\def\1rk#1{\cellcolor{color_1st}#1}
\def\2rk#1{\cellcolor{color_2nd}#1}
\def\3rk#1{\cellcolor{color_3rd}#1}
\usepackage{subcaption}
\usepackage{placeins}
\usepackage{float} % For [H] non-floating placement of algorithms
\usepackage[ruled]{algorithm2e} % For algorithms

\SetAlCapHSkip{0pt}

\def\paperID{*****} % *** Enter the Paper ID here
\def\confName{CVPR}

\def\confYear{2026}

\title{ForeSplat: Optimization-Aware Foresight for Feed-Forward 3D Gaussian Splatting}

\author{
Yuke Li$^{1,*}$,
Weihang Liu$^{1,2,*,\dag}$, 
Cheng Zhang$^{1,2,*}$,
Yuefeng Zhang$^{1,2,*}$,
Jiadi Cui$^{3}$,
Zixuan Wang$^{1}$,
\\
Junran Ding$^{1}$,
Haoyu Wu$^{1}$,
Yujiao Shi$^{1}$,
Jingyi Yu$^{1}$,
Xin Lou$^{1,2,\dag}$
\\
\text{$^{1}$ShanghaiTech University},
\text{$^{2}$GGU Technology Co., Ltd},
\text{$^{3}$Stereye}
}

\begin{document}

\twocolumn[{%
\maketitle
\renewcommand\twocolumn[1][]{#1}%
\includegraphics[width=1.0\linewidth]{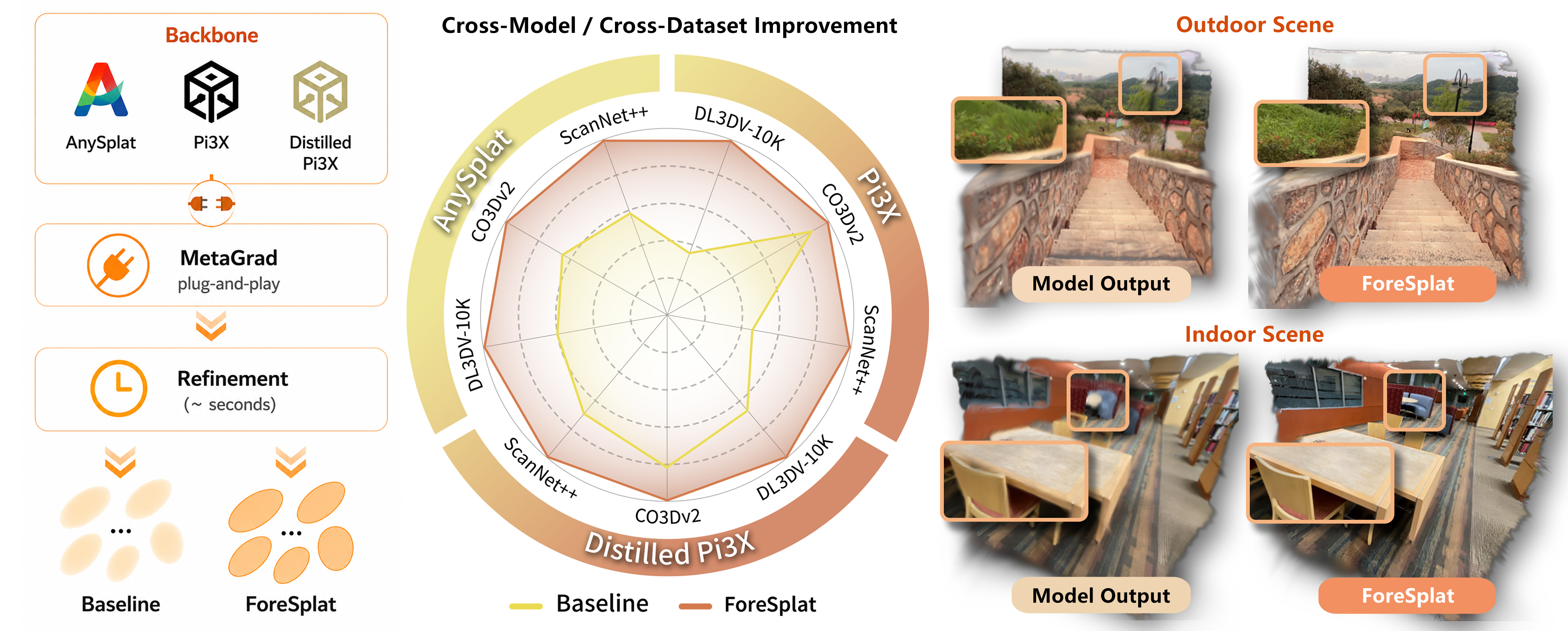}
% \vspace{-2em}
    \captionof{figure}{ForeSplat equips feed‑forward 3D Gaussian Splatting with an optimization‑aware training objective, making its predictions amenable to rapid per‑scene refinement.
    % At its heart is MetaGrad, a plug‑and‑play meta‑gradient training rule that can be integrated into any feed‑forward 3DGS architecture. 
    We showcase its generality by attaching it to diverse backbones, including AnySplat, Pi3X, and a distilled variant.
    The resulting post-optimization improvements are observed across multiple model families, datasets, and both indoor and outdoor scenes.}
\label{fig:teaser}
}]

\def\thefootnote{*}\footnotetext{Equal Contribution.}
\def\thefootnote{\dag}\footnotetext{Corresponding author.}
%% sec/0_abstract.tex
%% Abstract for ForeSplat (SIGGRAPH Asia 2026)

\begin{abstract}
    Feed‑forward 3D Gaussian Splatting (3DGS) models offer fast single‑pass reconstruction, but scaling them to match per‑scene optimization quality is fundamentally hindered by the scarcity of large‑scale 3D annotations. A practical compromise is predict‑then‑refine, where post‑prediction optimization compensates for the limited capacity of the feed‑forward network. However, standard feed‑forward 3DGS is trained solely for zero‑step rendering error, ignoring whether its output constitutes a good initialization for the downstream optimizer.
    We present ForeSplat, an optimization‑aware training framework that equips feed‑forward 3DGS models to produce initializations explicitly designed for rapid, effective refinement. By offloading part of the scene‑modeling burden to the optimizer, ForeSplat substantially reduces the capacity pressure on the feed‑forward model, making high‑quality reconstruction feasible even with compact networks. At its core is MetaGrad, a lightweight multi‑anchor meta‑gradient training rule that bypasses costly higher‑order differentiation through the 3DGS optimizer. MetaGrad unrolls a short inner‑loop refinement trajectory, samples anchor states, and back‑propagates aggregated first‑order gradients to the prediction head as a surrogate optimization‑aware signal. 
    This fine‑tuning adds no inference cost and enables high‑quality reconstruction within seconds after a few refinement steps.
    We instantiate ForeSplat on diverse backbones, including AnySplat, Pi3X, and a distilled variant tailored for edge deployment. Across all tested architectures, a ForeSplat‑trained initialization converges in fewer refinement steps and reaches a higher peak reconstruction quality than its vanilla counterpart, even fully converged. The framework consistently bridges the gap between amortized prediction and per‑scene optimization, establishing a practical path toward lightweight, high‑fidelity 3D reconstruction.

  % Feed-forward 3D Gaussian Splatting (3DGS) models such as AnySplat and Pi3X can predict scene representations in a single forward pass, but their outputs are not necessarily well conditioned for subsequent per-scene optimization. 
  % This limits their effectiveness in practical predict-then-refine pipelines, where the quality after a fixed refinement budget matters more than the zero-step rendering quality alone. 
  % We present \textbf{ForeSplat}, an optimization-aware training framework that teaches feed-forward 3DGS models to produce Gaussian initializations that are easier for the downstream optimizer to refine. 
  % At the core of ForeSplat is \textbf{MetaGrad}, a multi-anchor first-order meta-gradient training rule that samples short refinement trajectories and uses the losses observed along them to supervise the Gaussian prediction head. 
  % ForeSplat requires only a short fine-tuning stage, keeps the deployed network architecture unchanged, and introduces \emph{zero} additional inference cost. 
  % Instantiated on AnySplat, Pi3X, and Distill Pi3X, a distilled variant of Pi3X, ForeSplat consistently improves later-stage post-optimization quality across standard NVS metrics, reaches higher end-of-budget quality under matched refinement budgets, and requires fewer refinement steps to match the vanilla checkpoints at the same target quality. 
  % These results show that optimization-aware training provides a simple and effective bridge between feed-forward 3DGS prediction and per-scene 3DGS refinement.
\end{abstract}
%% sec/1_intro.tex
\section{Introduction}
\label{section:intro}
Feed-forward 3D Gaussian Splatting (3DGS) models~\cite{pixelsplat,MVSplat,jiang2025anysplat} reconstruct a scene from sparse views in a single forward pass, avoiding the high cost of per-scene optimization. 
Despite this efficiency, their one-shot rendering quality remains inferior to that of fully optimized per-scene reconstructions.
% largely because their model capacity is constrained by the limited 3D training data available.
A natural direction, therefore, is to scale up these feed-forward models until their output rivals per-scene optimization quality.
The core obstacle, however, is the widely recognized scarcity of large-scale, high-quality 3D scene annotations~\cite{MegaSynth}: unlike 2D vision, acquiring dense ground truth for diverse, unbounded scenes remains costly and often noisy. 
Consequently, feed-forward 3DGS models are trained on limited datasets, and further increases in model capacity yield diminishing returns as the supervisory signal saturates.

A practical compromise is \emph{predict-then-refine}: use the feed-forward output as an initialization and apply a modest budget of per-scene optimization~\cite{jiang2025anysplat, InstantSplat, MVSGaussian}. This shifts the central requirement from single-pass quality to providing an initialization that is well conditioned for rapid refinement. Yet existing feed-forward 3DGS training does not reflect this deployment objective. 
Let $G_0=f_\Theta(\mathcal{I})$ be the Gaussian scene predicted from inputs $\mathcal{I}$. Standard training minimizes a loss on $G_0$, whereas a predict-then-refine pipeline ultimately evaluates the refined state $G_K$ after $K$ steps of optimization. 
A prediction with low zero-step error is not necessarily an optimal starting point: it may lie in a region of the 3DGS parameter space where convergence is slow, local gradients are uninformative, 
% early steps are dominated by structural cleanup rather than appearance refinement, 
or the optimization plateaus prematurely under a fixed budget.

This mismatch motivates optimization-aware training: feed-forward 3DGS should be evaluated not only by its immediate rendering quality, but by the quality attainable after a downstream refinement budget. In light of the severe scarcity of annotated 3D data, we reframe the problem as learning to optimize, equipping the feed-forward model with the meta-learning capability to produce initializations that are amenable to rapid refinement, rather than to directly predict the final output.
Directly optimizing this meta‑objective, however, requires back‑propagating through a multi‑step 3DGS refinement trajectory where the Gaussian scene typically contains tens of millions of parameters. Consequently, even a modest number of inner steps incurs prohibitive memory and involves second‑order derivatives whose Hessian matrices are far too large to compute or store.

We present \textbf{ForeSplat}, an optimization-aware training framework for feed-forward 3DGS. ForeSplat preserves the inference-time architecture of the host model but retrains its Gaussian prediction head to produce initializations that the downstream optimizer can rapidly improve. 
We formulate this task as a meta-learning problem, where the objective is to learn an initial set of parameters that, following a brief refinement process, yields low loss. Directly back-propagating through the multi-step 3DGS optimizer would entail expensive second-order derivatives. Furthermore, because the training iterations often exceed hundreds or even thousands, unrolling the optimization process, as in a conventional meta-learning framework, leads to increased training time and instability~\cite{Unrolled_meta1, Unrolled_meta2, Unrolled_meta3}.
% As the training iterations can exceeds hundreds or even thousands, unrolling the optimization process as a conventional meta-learning task results in increased training time and instability. 

To circumvent this, we introduce \textbf{MetaGrad}, a lightweight meta-gradient training rule that avoids high-order derivatives. 
During training, MetaGrad unrolls a short per-scene refinement trajectory from the feed-forward prediction, samples several anchor states along that trajectory, and evaluates photometric losses at these anchors to form an aggregate meta-objective $L_{\mathrm{meta}}$. To propagate the supervision back to the Gaussian prediction head, we differentiate $L_{\mathrm{meta}}$ with respect to the sampled anchor states, accumulate the resulting gradients, and use their sum as a surrogate error signal for the initial prediction $G_0$. This surrogate signal is then back-propagated through the feed-forward Gaussian head. In this way, MetaGrad encourages the network to predict initialization-friendly Gaussian parameters that are better suited for subsequent refinement, while avoiding full higher-order differentiation through the optimizer.

We instantiate ForeSplat on multiple feed-forward backbones, including AnySplat~\cite{jiang2025anysplat} and Pi3X~\cite{wang2025pi3}. By offloading part of the scene modeling burden to the downstream optimizer, ForeSplat substantially reduces the capacity pressure on the feed-forward model. This makes it feasible to learn scene prediction with smaller networks, opening up new possibilities for deploying 3D reconstruction on edge devices and in resource-constrained settings. To demonstrate its effectiveness in this lightweight regime, we further introduce a distilled variant of Pi3X. 

Across all tested backbones, a ForeSplat-trained initialization converges more rapidly and to a higher peak reconstruction quality than the vanilla baseline can attain.
In summary, our contributions are as follows:
\begin{itemize}
    \item \textit{Optimization-aware training for feed-forward 3DGS.}
    We identify a training–deployment mismatch: existing models are trained for raw prediction error, while predict-then-refine pipelines depend on post-refinement quality.
    The proposed ForeSplat addresses it by meta-training the network to minimize post-optimization loss without requiring larger 3D datasets.
    \item \textit{Multi-anchor meta-gradient rule.}
    We propose MetaGrad, which bypasses higher-order differentiation through the 3DGS optimizer by aggregating first-order gradients sampled at multiple anchor states along the inner-loop trajectory. The multi-anchor design stabilizes the meta-objective without incurring second-order cost.
    \item \textit{Plug-in compatibility across backbones.}
    Applied to multiple feed-forward backbones (AnySplat, Pi3X, and a distilled compact Pi3X), ForeSplat consistently improves post-refinement quality under a fixed optimization budget without architectural changes or inference-time overhead, paving the way for lightweight deployment of 3D models.
\end{itemize}

\section{Related Work}
\label{section:related}

\subsection{Optimization-based Novel View Synthesis}
Optimization-based methods have long dominated novel view synthesis.
The central bottleneck in this family is the tension between rendering quality and deployment cost. These approaches, exemplified by Neural Radiance Field (NeRF)~\cite{NeRF,CoQ,CoARF,mipnerf360,NGP} and 3D Gaussian Splatting (3DGS)~\cite{3DGS}, achieve impressive fidelity by iteratively fitting a scene representation to input images using stochastic optimizers such as Adam~\cite{kingma2014adam}. However, this process demands two expensive prerequisites: (i) accurate camera poses, conventionally obtained via Structure-from-Motion (SfM)~\cite{SfM1, SfM2} tools like COLMAP, and (ii) minutes to hours of per-scene optimization~\cite{duplexGS,CityGo}. Even recent work that jointly optimizes poses and scene representation~\cite{lin2021barf, fu2024colmapfree} does not eliminate these costs; it merely reduces them incrementally. As a result, deploying optimization-based methods in unconstrained, real-time settings remains impractical. This gap has motivated a parallel line of research that replaces per-scene optimization with amortized feed-forward inference.

\subsection{Feed-forward 3D reconstruction}
\paragraph{Generalizable 3D foundation models.}
A parallel line of work trains large-scale models to predict scenes directly from images without per-scene optimization. DUSt3R~\cite{DUSt3R} and its successors MAST3R~\cite{MAST3R} and Fast3R~\cite{Fast3R} jointly estimate depth and camera poses from unordered image pairs or collections, producing dense point clouds. VGGT~\cite{wang2025vggt, PanoVGGT} and Pi3X~\cite{wang2025pi3} further extend this to multi-view pose estimation and point tracking. While these models excel at geometric reconstruction, they do not directly support novel view synthesis because they lack a differentiable renderer for radiance fields.

\paragraph{Feed-forward 3D Gaussian Splatting.}
To enable instant novel view synthesis, several works combine feed-forward prediction with 3DGS, building on a longer line of generalizable neural rendering pioneered by pixelNeRF~\cite{yu2021pixelnerf} and large reconstruction models such as LRM~\cite{hong2023lrm}. pixelSplat~\cite{pixelsplat} and MVSplat~\cite{MVSplat} predict per-pixel Gaussians from sparse, calibrated views using cost volumes. LatentSplat~\cite{latentSplat} models variational Gaussians in a latent space. Object-level variants such as Splatter Image~\cite{splatterImage}, GS-LRM~\cite{GSLRM}, LGM~\cite{LGM}, GRM~\cite{GRM}, and AGG~\cite{AGG} scale transformer backbones to Gaussian prediction. More recent work pushes toward many-view, pose-free, or in-the-wild scene-level settings: Flash3D~\cite{Flash3D} builds on a monocular depth foundation model; DepthSplat~\cite{DepthSplat} couples feed-forward Gaussians with multi-view depth; FreeSplat~\cite{FreeSplat} aggregates redundancy across long indoor sequences; NoPoSplat~\cite{NoPoSplat} eliminates known poses; Long-LRM~\cite{LongLRM} scales to 32 input views. AnySplat~\cite{jiang2025anysplat} removes the need for camera poses by jointly predicting Gaussians and camera parameters from unposed images. YoNoSplat~\cite{YoNoSplat} handles arbitrary numbers of views and can optionally use ground-truth poses when available. EcoSplat~\cite{EcoSplat} introduces controllable primitive count via importance-aware training. Across all these methods, the network is supervised to produce the final radiance field in one shot; when the prediction is used as-is, a persistent quality gap to a full per-scene 3DGS optimization remains.

% To enable instant novel view synthesis, several works combine feed-forward prediction with 3DGS. pixelSplat~\cite{pixelsplat} and MVSplat~\cite{MVSplat} predict per-pixel Gaussians from sparse, calibrated views using cost volumes. AnySplat removes the need for camera poses by jointly predicting Gaussians and camera parameters from unposed images. YoNoSplat~\cite{YoNoSplat} handles arbitrary numbers of views and can optionally use ground-truth poses when available. EcoSplat~\cite{EcoSplat} introduces controllable primitive count via importance-aware training. Despite their speed, all these methods are trained to minimize the rendering error of their raw prediction—the objective is instantaneous quality, not the behavior under subsequent optimization.

\paragraph{Predict-then-refine and the training--deployment mismatch.}
A natural extension is to use feed-forward prediction as an initializer for a few steps of per-scene refinement. This predict-then-refine strategy has been adopted in recent pipelines: AnySplat~\cite{jiang2025anysplat} itself applies post-optimization to boost quality, and dedicated works such as InstantSplat~\cite{InstantSplat}, Splatt3R~\cite{smart2024splatt3r}, and MVSGaussian~\cite{MVSGaussian} show that a modest number of refinement steps closes much of the gap between feed-forward and fully optimized reconstructions.
However, a critical mismatch remains: the feed-forward network is still trained solely to minimize the error of its own prediction. It never receives supervision from the quality achieved after refinement. Consequently, the predicted Gaussians may lie in regions of the parameter space where refinement converges slowly, responds poorly to local gradients, or plateaus early. The network is not encouraged to produce an initialization that is easy to improve. Instead, it is only rewarded for being good from the start. This disconnect between training objective and deployment goal motivates our work.

\subsection{Meta-Learning}
Meta-learning extracts transferable knowledge across tasks to enable rapid adaptation from limited data. Optimization-based methods such as MAML~\cite{finn2017model} and Reptile~\cite{nichol2018first} find a weight initialization from which a few gradient steps suffice for new tasks, with subsequent works providing implicit-gradient variants~\cite{rajeswaran2019meta} and validating first-order training with adaptive optimizers~\cite{antoniou2018train}; a parallel line learns the optimizer itself~\cite{andrychowicz2016learning}. Tancik et al.~\cite{tancik2021learned} applied this idea to NeRF, MetaSDF~\cite{sitzmann2020metasdf} extended it to signed distance fields, and FewShotNeRF~\cite{FewShotNeRF} extended it to category-specific scene adaptation. Our work, ForeSplat, applies a first-order meta-gradient to train a feed-forward 3DGS model. This encourages initializations that achieve higher quality after a shorter per-scene refinement. Crucially, this approach incurs no inference-time overhead, thereby paving the way for lightweight deployment of 3D models.

\section{Method}
\label{section:method}
\begin{figure*}[t]
    \centering
    \includegraphics[width=\textwidth]{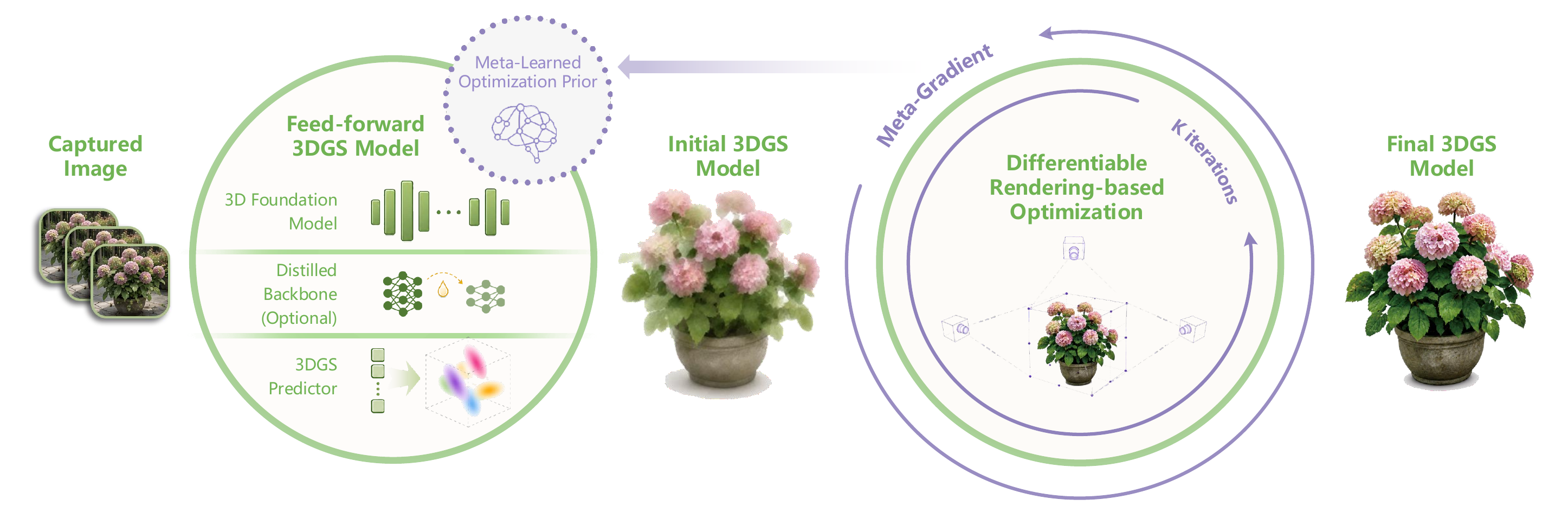}
    \caption{Overview of ForeSplat. Given a set of uncalibrated input images, a feed‑forward 3DGS model predicts initial Gaussians. ForeSplat then unrolls a short refinement trajectory, samples anchor states, and back‑propagates the resulting first‑order meta‑gradients through the Gaussian prediction head. This produces optimization‑aware initializations without altering the inference‑time pipeline.
    % Overview of ForeSplat. Given a set of uncalibrated images, a feed-forward 3DGS model predicts initial Gaussians. ForeSplat then runs short refinement rollouts, records meta-gradients along the optimization trajectory, and sends them back to the Gaussian head, yielding optimization-aware initializations without changing the inference-time pipeline.
    }
    % \Description{pipeline}
    \label{fig:pipeline}
\end{figure*}

% We propose \emph{ForeSplat}, a plug-and-play optimization-aware training framework that turns feed-forward 3D Gaussian Splatting model into a producer of \emph{optimization-aware} initial Gaussians whose subsequent per-scene refinement converges faster and to a higher quality ceiling.
We propose \emph{ForeSplat}, a plug‑and‑play optimization‑aware training framework that turns a feed‑forward 3DGS model into an initialization generator whose predictions converge faster and to a higher peak quality under per‑scene refinement.
As shown in Fig.~\ref{fig:pipeline}, ForeSplat first predicts an initial set of Gaussians from uncalibrated images and then unrolls a short per‑scene refinement trajectory from this prediction. The meta‑gradients back‑propagated along the unrolled steps serve as the supervisory signal for the Gaussian prediction head.
Crucially, ForeSplat leaves the inference‑time architecture of the host network untouched; the change is confined to the training objective and the gradient flow induced by the unrolled optimizer.
In this section, we first establish the necessary notation (\cref{section:method:prelim}) and formalize the optimization‑aware learning problem (\cref{section:method:problem}). 
Building on this, we derive the second‑order meta‑gradient and its tractable first‑order approximation for 3DGS (\cref{section:method:fomaml}), and finally introduce \emph{MetaGrad}, the multi‑anchor first‑order training rule at the core of ForeSplat (\cref{section:method:metagrad}).
% As illustrated in \cref{fig:pipeline}, ForeSplat first predicts initial Gaussians from a set of uncalibrated images, then performs short refinement rollouts during training and feeds the resulting meta-gradients back to the Gaussian head.
% The inference graph of the host network is left untouched; ForeSplat modifies only the training-time update rule and its associated gradient flow.
% We first set up notation (\cref{section:method:prelim}) and formalize the optimization-aware learning problem (\cref{section:method:problem}).
% We then derive the second-order meta-gradient and its tractable first-order approximation for 3DGS (\cref{section:method:fomaml}), and present \emph{MetaGrad}, the multi-anchor first-order training rule at the core of ForeSplat (\cref{section:method:metagrad}).

\subsection{Preliminaries}
\label{section:method:prelim}

\paragraph{Per-scene Optimization 3DGS}
Given a set of input images, a standard 3DGS pipeline first recovers camera poses and an initial sparse point cloud, typically using SfM or learning‑based alternatives such as MASt3R and VGGT. These points serve as the centers $\mu_i$ of a set of anisotropic Gaussian primitives $G = \{(\mu_i, s_i, q_i, \alpha_i, c_i)\}_{i=1}^{|G|}$, where $s_i \in \mathbb{R}^3$ is the scale, $q_i$ a rotation quaternion, $\alpha_i$ the opacity, and $c_i$ a view‑dependent color encoded via spherical harmonics~\cite{3DGS}. The scene representation is then refined by repeatedly rendering $G$ to the training viewpoints and minimizing a photometric loss $L_A(G)$ using the Adam optimizer for tens of thousands of iterations. The loss typically combines an $L_1$ term with a D‑SSIM term and is often augmented with opacity and scale regularizers. This optimization may also include adaptive control strategies such as densification, cloning, splitting, and pruning.
% the scene is reconstructed by repeatedly rendering the current Gaussian set $G$ to the training viewpoints and minimizing a photometric loss $L_A(G)$ using the Adam optimizer for tens of thousands of iterations. The loss is typically a combination of $L_1$ and D-SSIM, augmented with opacity and scale regularizers.

\paragraph{Feed-forward 3DGS}
A feed-forward model $f_\Theta : \mathcal{I} \mapsto G_0$ predicts a 3D Gaussian scene $G_0 = f_\Theta(\mathcal{I})$ from input images $\mathcal{I}$ in a single forward pass. The network parameters $\Theta$ typically consist of a 3D backbone and a Gaussian prediction head.

% \paragraph{Per-scene post-optimization}
% Given the feed-forward prediction $G_0$, we may further refine it with $K$ steps of standard 3DGS optimization, producing a trajectory $G_0 \to G_1 \to \cdots \to G_K$.
% We denote this operator $\mathrm{Opt}_K(G_0) := G_K$.
% In our experiments, this post-optimization uses a fixed Gaussian topology: structural operations such as densification, cloning, splitting, and pruning are disabled, so the inner loop optimizes only the attributes of the predicted Gaussians.
% For analytical purposes we write a generic inner step as
% \begin{equation}
% \label{eq:inner_step}
% G_{k+1} = G_k - \eta \, \nabla_{G_k} L_A(G_k),
% \end{equation}
% where $\eta$ is the inner learning rate.
% \cref{eq:inner_step} uses SGD as a tractable proxy; in practice we use Adam, and the analytical structure that follows remains valid up to a first-order approximation~\cite{antoniou2018train}.

\subsection{Problem Formulation: Optimization-Aware Initialization}
\label{section:method:problem}
We combine feed‑forward 3DGS with per‑scene post‑optimization to achieve lightweight, high‑quality reconstruction within seconds, without requiring explicit 3D supervision (e.g., ground-truth point clouds).
Starting from the feed‑forward prediction $G_0$, we apply $K$ steps of refinement, yielding a trajectory $G_0 \to G_1 \to \cdots \to G_K$; we denote this operator by $\mathcal{R}_K(G_0) := G_K$. 
In our experiments, the post‑optimization keeps the number of Gaussian primitives fixed: structural operations such as densification, cloning, splitting, and pruning are disabled, so the inner loop updates only the attributes of the predicted Gaussians. For analytical purposes, we write a generic inner step as
\begin{equation}
\label{eq:inner_step}
G_{k+1} = G_k - \eta \, \nabla_{G_k} L_A(G_k),
\end{equation}
where $\eta$ is the inner learning rate. Eq.~\ref{eq:inner_step} uses SGD as a tractable proxy; in practice we use Adam, and the analytical structure that follows remains valid~\cite{antoniou2018train}.

The conventional training objective for feed‑forward 3DGS minimizes the immediate rendering loss of the predicted Gaussians:
\begin{equation}
\label{eq:vanilla_obj}
\min_\Theta \; \mathbb{E}_{\mathcal{I}}\!\left[\, L_A\!\big(f_\Theta(\mathcal{I})\big)\,\right].
\end{equation}
In our design, however, the feed‑forward prediction serves only as a starting point for downstream per‑scene refinement; the final quality after refinement is what matters. We therefore optimize the post‑optimization loss directly:
\begin{equation}
\label{eq:meta_obj}
\min_\Theta \; \mathbb{E}_{\mathcal{I}}\!\left[\, L_A\!\big(\mathcal{R}_K(f_\Theta(\mathcal{I}))\big)\,\right].
\end{equation}
This reformulates feed‑forward 3DGS training as a learning‑to‑optimize problem: we seek an initialization $G_0=f_\Theta(\mathcal{I})$ that is well aligned with the loss landscape navigated by the downstream optimizer. Eq.~\ref{eq:meta_obj} is the canonical Model‑Agnostic Meta‑Learning (MAML) objective~\cite{finn2017model} instantiated for the feed‑forward 3DGS pipeline.

% The conventional feed-forward 3DGS training objective minimizes the \emph{immediate} rendering loss of the feed-forward output:
% \begin{equation}
% \label{eq:vanilla_obj}
% \min_\Theta \; \mathbb{E}_{\mathcal{I}}\!\left[\, L_A\!\big(f_\Theta(\mathcal{I})\big)\,\right].
% \end{equation}
% However, in many deployment scenarios the feed-forward prediction is only the \emph{starting point} of a downstream per-scene refinement; what ultimately matters is the quality \emph{after} that refinement.
% We therefore propose to optimize the \emph{post-optimization} loss directly:
% \begin{equation}
% \label{eq:meta_obj}
% \min_\Theta \; \mathbb{E}_{\mathcal{I}}\!\left[\, L_A\!\big(\mathcal{R}_K(f_\Theta(\mathcal{I}))\big)\,\right].
% \end{equation}
% This recasts FF-3DGS training as a \emph{learning-to-optimize} problem: we seek an initialization $G_0=f_\Theta(\mathcal{I})$ that is well-aligned with the loss landscape navigated by Adam during per-scene refinement.
% \cref{eq:meta_obj} is the canonical Model-Agnostic Meta-Learning (MAML) objective~\cite{finn2017model} instantiated on the FF-3DGS pipeline.

\subsection{Meta-Gradient via MAML and its First-Order Approximation}
\label{section:method:fomaml}

\subsubsection{Second-order meta-gradient}
\label{section:method:fomaml:second}
To minimize Eq.~\ref{eq:meta_obj} by gradient descent, we require the derivative $\nabla_\Theta L_A(G_K)$.
Applying the chain rule along the inner trajectory Eq.~\ref{eq:inner_step} rewrites this as
\begin{equation}
\label{eq:chain}
\nabla_{G_0} L_A(G_K)
= \nabla_{G_K} L_A(G_K) \cdot \prod_{k=0}^{K-1} \frac{\partial G_{k+1}}{\partial G_k}.
\end{equation}
From Eq.~\ref{eq:inner_step} we obtain
\begin{equation}
    \frac{\partial G_{k+1}}{\partial G_k} = \mathbf{I} - \eta \,\nabla^2_{G_k} L_A(G_k),
\end{equation}
where $\mathbf{I}$ is the identity matrix and $\nabla^2_{G_k} L_A$ denotes the Hessian of $L_A$ at $G_k$.
Substituting this term yields the full \emph{second-order} meta-gradient:
\begin{equation}
\label{eq:second_order}
\nabla_{G_0} L_A(G_K)
= \nabla_{G_K} L_A(G_K) \cdot \prod_{k=K-1}^{0}\!\Big(\mathbf{I} - \eta \,\nabla^2_{G_k} L_A(G_k)\Big).
\end{equation}
Finally, the gradient is back-propagated to the network weights via
\begin{equation}
\label{eq:to_theta}
\nabla_\Theta L_A(G_K) = \nabla_{G_0} L_A(G_K) \cdot \frac{\partial f_\Theta(\mathcal{I})}{\partial \Theta}.
\end{equation}

\subsubsection{Intractability of the Second‑Order Meta‑Gradient}
\label{section:method:fomaml:intractable}
While training with the full second‑order meta‑gradient appears conceptually attractive, two fundamental obstacles render it intractable in the 3DGS setting.

\textbf{(i) Hessian dimension.}
The Gaussian state $G$ typically contains $10^7$ to $10^8$ parameters. Consequently, the Hessian $\nabla^2_{G_k} L_A \in \mathbb{R}^{|G|\times|G|}$ can be neither explicitly formed nor directly multiplied. Even Hessian–vector products through the splatting rasterizer are prohibitively slow.

\textbf{(ii) Memory of unrolled back‑propagation.}
Differentiating through $K$ inner Adam steps requires caching all intermediate states $\{G_k, \text{Adam moments}\}_{k=0}^{K}$, which adds $\mathcal{O}(K)$ memory on top of an already memory‑intensive rasterization graph.
As illustrated in~\cref{fig:metagrad}, native $K$‑step unrolling for second‑order meta‑learning demands a full sequential backward pass through the entire optimization trajectory, whereas our sparse‑anchor alternative avoids this costly unrolled back‑propagation.

% Two factors make \cref{eq:second_order} infeasible in the 3DGS regime.
% \textbf{(i) Hessian dimension.}
% The Gaussian state $G$ has dimension $|G|\!\sim\!10^7$--$10^8$, so the Hessian $\nabla^2_{G_k} L_A \in \mathbb{R}^{|G|\times|G|}$ can neither be materialized nor multiplied; even Hessian--vector products through the splatting rasterizer are prohibitively slow.
% \textbf{(ii) Memory of unrolled backpropagation.}
% Differentiating through $K$ inner Adam steps requires retaining the entire intermediate state $\{G_k, \text{Adam moments}\}_{k=0}^{K}$, incurring $\mathcal{O}(K)$ memory on top of an already memory-bound rasterization graph.
% This difficulty is illustrated in \cref{fig:metagrad}: native $K$-step unrolling for second-order meta-learning requires sequential back-propagation through the full optimization trajectory, whereas our sparse-anchor alternative avoids this full unrolled backward pass.

\subsubsection{First-order approximation (FOMAML)}
\label{section:method:fomaml:fomaml}
Following Finn et al.~\cite{finn2017model}, we drop all second-order terms by approximating $\partial G_{k+1}/\partial G_k \approx \mathbf{I}$, which collapses Eq.~\ref{eq:second_order} to
\begin{equation}
\label{eq:fomaml}
\nabla_{G_0} L_A(G_K) \;\approx\; \nabla_{G_K} L_A(G_K),
\end{equation}
i.e.\ the photometric gradient evaluated at the \emph{post-optimization} state $G_K$ is used as a \emph{surrogate gradient} on the \emph{pre-optimization} state $G_0$, and from there propagated to $\Theta$ through a single backward pass:
\begin{equation}
\label{eq:fomaml_theta}
\nabla_\Theta L_A(G_K) \;\approx\; \nabla_{G_K} L_A(G_K) \cdot \frac{\partial f_\Theta(\mathcal{I})}{\partial \Theta}.
\end{equation}
The intuition is that, after $K$ inner steps, the descent direction at $G_K$ remains a meaningful supervisory signal for the initialization $G_0$.
This approximation is known to perform competitively with the full second-order variant in practice~\cite{finn2017model,nichol2018first}, while reducing memory from $\mathcal{O}(K)$ to $\mathcal{O}(1)$ in the inner loop.

\subsection{MetaGrad: Multi-Anchor FOMAML along the Optimization Trajectory}
\label{section:method:metagrad}

\begin{figure}
    \centering
    \includegraphics[width=1.0\linewidth]{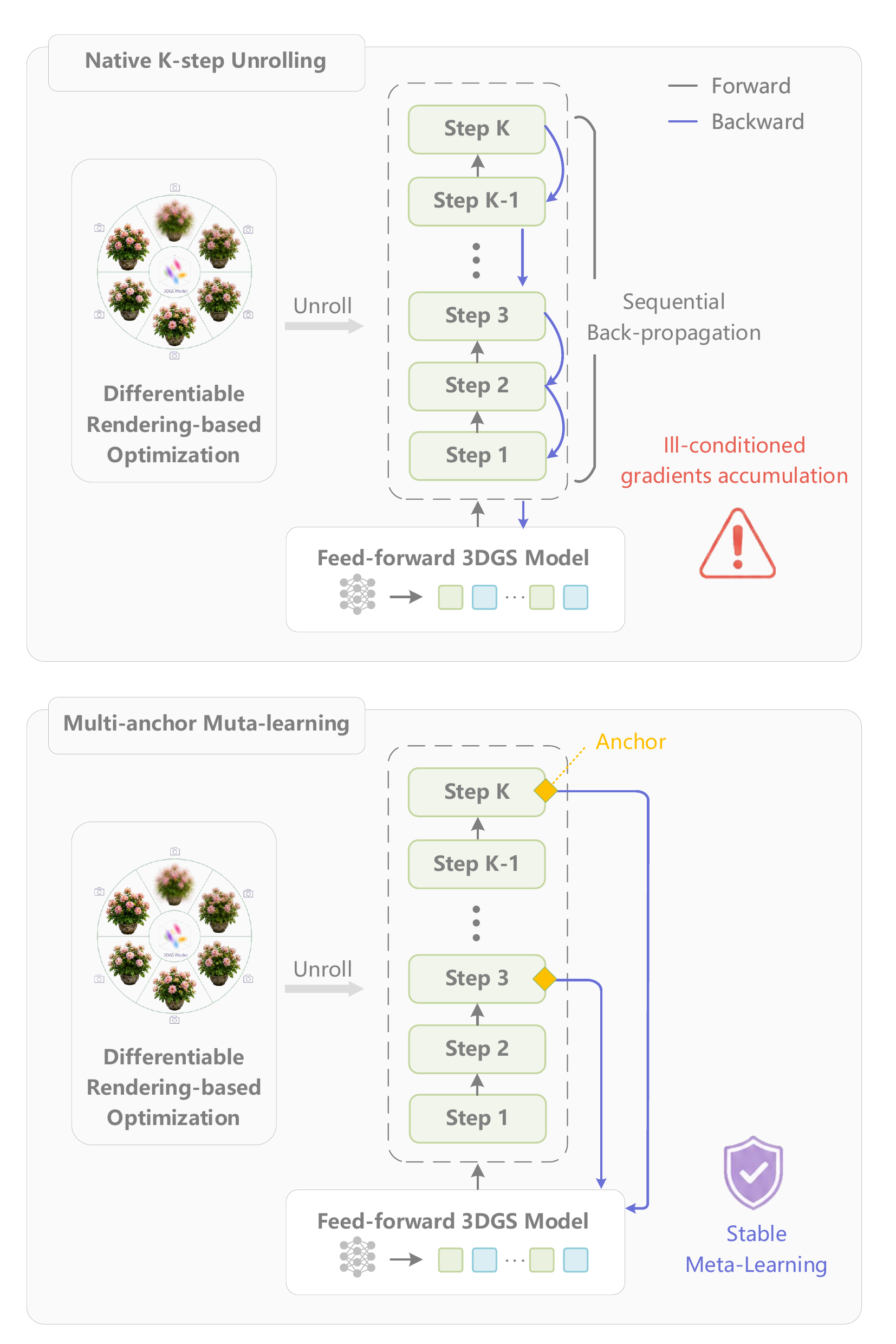}
    \caption{Comparison between native $K$-step unrolling and MetaGrad. Top: native $K$-step unrolling for second-order meta-learning differentiates through the full refinement trajectory, requiring sequential back-propagation across all inner steps and accumulating ill-conditioned gradients. Bottom: MetaGrad samples sparse anchor states along the trajectory and routes their first-order surrogate gradients directly back to the feed-forward 3DGS model, enabling more stable meta-learning.}
    % \Description{metagrad}
    \label{fig:metagrad}
\end{figure}

\paragraph{Motivation}
Single-endpoint FOMAML supervises the network solely at a fixed, pre‑defined optimization horizon $K$, leaving the intermediate states $G_1,\dots,G_{K-1}$ unconstrained.
In practice, this can make the learned initialization sensitive to the chosen $K$, leading to degraded performance when the number of refinement steps at deployment deviates substantially from the training‑time $K$.

An alternative is to supervise every step along the trajectory. Reptile~\cite{nichol2018first} takes this approach by accumulating the overall parameter displacement as a surrogate update direction:
\begin{equation}
\label{eq:reptile_theta}
\hat{g}^{\mathrm{reptile}}_\Theta
\;=\;
\left(G_0 - G_{K_{\mathrm{rand}}}\right)
\;\cdot\;
\frac{\partial f_\Theta(\mathcal{I})}{\partial \Theta}.
\end{equation}
However, in 3DGS post‑optimization the inner trajectory can be long, and densely supervising every step, as illustrated in Fig.~\ref{fig:metagrad}, introduces noise and training instability.

% Another meta-learning method Reptile~\cite{nichol2018first} supervises the entire $K$ steps using 
% \begin{equation}
% \label{eq:reptile_theta}
% \hat{g}^{\mathrm{rep}}_\Theta
% \;=\;
% \left(G_0 - G_{K_{\mathrm{rand}}}\right)
% \;\cdot\;
% \frac{\partial f_\Theta(\mathcal{I})}{\partial \Theta}.
% \end{equation}
% which supervise the summation of gradients from all steps. As has been mentioned above, the training trajectory in 3DGS optimization scenarios could be long where, as shown in Fig.~\ref{fig:metagrad}, recording gradients of every iteration can cause instability.

\paragraph{Trajectory sampling}
We address this by supervising multiple horizons along the optimization trajectory and aggregating their surrogate gradients via equal-weight multi-horizon averaging.
Let $K_{\max}$ denote the maximum unroll horizon used during training and $\Delta$ the anchor stride (we use $\Delta = 40$).
At each training iteration we sample
\begin{equation}
\label{eq:k_rand}
K_{\mathrm{rand}} \sim \mathcal{U}\{K_{\min}, \dots, K_{\max}\},
\end{equation}
and roll out $K_{\mathrm{rand}}$ Adam steps of per-scene refinement, caching only the sparse anchor set
\begin{equation}
\label{eq:anchor_set}
\begin{gathered}
\mathcal{K} = \{\, k \in \mathbb{Z} \mid 1 \le k \le K_{\mathrm{rand}},\ k \bmod \Delta = 0 \,\}, \\
N_{\mathrm{anchor}} = |\mathcal{K}|.
\end{gathered}
\end{equation}
Randomizing the horizon prevents overfitting to any single $K$ and exposes the network to a curriculum of refinement depths.

\paragraph{Multi-anchor loss aggregate}
We define the post-optimization-aware scalar anchor-loss aggregate, denoted $L_{\mathrm{meta}}$ (MetaLoss), as the equal-weight average of the photometric losses at the sampled anchors,
\begin{equation}
\label{eq:l_meta}
L_{\mathrm{meta}} \;=\; \frac{1}{N_{\mathrm{anchor}}} \sum_{k \in \mathcal{K}} L_A(G_k),
\end{equation}
% and combine it with the standard immediate-rendering loss to form the total ForeSplat training objective:
For exposition, we write the combined outer objective as
\begin{equation}
\label{eq:total_loss}
\mathcal{L}(\Theta) \;=\; \lambda\, L_A(G_0) \;+\; (1-\lambda) \, L_{\mathrm{meta}},
\qquad \lambda \in [0,1].
\end{equation}
Eq.~\ref{eq:total_loss} is shorthand for the outer-loop weighting only: in implementation, $L_A(G_0)$ is back-propagated directly, whereas $L_{\mathrm{meta}}$ contributes through the detached surrogate meta-gradient estimator described below rather than through end-to-end differentiation of the inner trajectory.

\paragraph{Equal-weight aggregation and sparse anchoring}
Three properties of Eq.~\ref{eq:l_meta} deserve comment.
\textbf{(i)~Equal weighting across anchors.}
Each $G_k$ corresponds to a legitimate horizon at which a user might stop refinement at deployment; absent a prior on the deployment horizon, we have no reason to favor any particular anchor and therefore weight them equally.
\textbf{(ii)~Sparse sampling with stride $\Delta = 40$.}
Consecutive Adam steps move $G_k$ by an amount that is small relative to its own scale, so $\nabla_{G_k} L_A(G_k)$ and $\nabla_{G_{k+1}} L_A(G_{k+1})$ are highly correlated and the marginal supervisory information of including every step is limited.
Meanwhile, every additional anchor incurs one extra photometric backward pass, so densely anchoring the trajectory significantly slows training.
\textbf{(iii)~Normalization by $N_{\mathrm{anchor}}$.}
The explicit $1/N_{\mathrm{anchor}}$ factor keeps the overall magnitude of $L_{\mathrm{meta}}$ invariant to the sampled $K_{\mathrm{rand}}$, so the optimal $\lambda$ in \cref{eq:total_loss} does not drift across iterations and hyper-parameter tuning is simplified.

\paragraph{Meta-gradient estimator}
% Differentiating \cref{eq:l_meta} term by term and applying the FoMAML approximation of \cref{eq:fomaml} to each anchor yields
MetaGrad does not back-propagate through Eq.~\ref{eq:l_meta} as a fully differentiable scalar objective. Instead, it computes first-order photometric gradients on detached anchor states, averages them, and attaches the resulting surrogate gradient to $G_0$. Applying the FOMAML approximation of \cref{eq:fomaml} to each anchor yields the surrogate meta-gradient estimator

% \begin{equation}
% \label{eq:meta_grad}
% \nabla_\Theta L_{\mathrm{meta}}
% \;\approx\;
% \left(\frac{1}{N_{\mathrm{anchor}}}\sum_{k \in \mathcal{K}} \nabla_{G_k} L_A(G_k)\right)
% \;\cdot\;
% \frac{\partial f_\Theta(\mathcal{I})}{\partial \Theta}.
% \end{equation}
\begin{multline}
\label{eq:meta_grad}
\hat{g}^{\mathrm{meta}}_\Theta
\;=\;
\left(\frac{1}{N_{\mathrm{anchor}}}\sum_{k \in \mathcal{K}} \nabla_{G_k} L_A(G_k)\right)
\;\cdot\;
\frac{\partial f_\Theta(\mathcal{I})}{\partial \Theta}
\\
\;\approx\; \nabla_\Theta L_{\mathrm{meta}}.
\end{multline}
The bracketed term is computed without back-propagation through the inner trajectory: each $\nabla_{G_k} L_A(G_k)$ is an independent first-order gradient on a detached tensor, and their average is then attached to $G_0$.

\section{Experiments}
\label{section:exp}

\subsection{Experimental Setup}
\label{section:exp:setup}
\label{sec:setup}

\paragraph{Datasets.}

% \begin{table}[t]
%     \centering
%     \small
%     \caption{Training datasets and sampling weights. At each
%     iteration, one dataset is drawn according to the listed weights.
%     % Real world, high diversity captures dominate the mix, while
%     % synthetic datasets fill in outdoor, city scale, and wide baseline
%     % regimes.
%     }
%     \setlength{\tabcolsep}{4pt}
%     \begin{tabular*}{\linewidth}{@{\extracolsep{\fill}} l c c @{}}
%         \toprule
%         Dataset & Type & Weight (\%) \\
%         \midrule
%         DL3DV-10K~\cite{ling2024dl3dv}           & Real             & 25 \\
%         CO3Dv2~\cite{reizenstein2021co3d}        & Real             & 20 \\
%         ARKitScenes~\cite{baruch2021arkitscenes} & Real             & 15 \\
%         WildRGB-D~\cite{xia2024wildrgbd}         & Real             & 10 \\
%         TartanAir~\cite{wang2020tartanair}       & Synthetic        & 8  \\
%         GTA-SfM~\cite{wang2020gtasfm}            & Synthetic        & 7  \\
%         ScanNet++~\cite{yeshwanth2023scannetpp}  & Real             & 5  \\
%         MatrixCity~\cite{li2023matrixcity}       & Synthetic        & 5  \\
%         BlendedMVS~\cite{yao2020blendedmvs}      & Real             & 5  \\
%         \bottomrule
%     \end{tabular*}
%     \label{tab:training_datasets}
%     \Description{dataset}
% \end{table}

Following standard practice in large-scale feed-forward 3D reconstruction models~\cite{jiang2025anysplat,wang2025pi3}, we train on a heterogeneous mixture of nine public multi-view datasets, including DL3DV-10K~\cite{ling2024dl3dv}, CO3Dv2~\cite{reizenstein2021co3d}, ARKitScenes~\cite{baruch2021arkitscenes}, WildRGB-D~\cite{xia2024wildrgbd}, TartanAir~\cite{wang2020tartanair}, GTA-SfM~\cite{wang2020gtasfm}, ScanNet++~\cite{yeshwanth2023scannetpp}, MatrixCity~\cite{li2023matrixcity}, and BlendedMVS~\cite{yao2020blendedmvs}. This combined dataset provides extensive coverage of object-centric, indoor, outdoor, and city-scale scenes, encompassing both synthetic and real-world scenarios. 
The input resolution follows the pretraining regime of each host backbone.
For Pi3X and Distill Pi3X, following the training setup of the original Pi3X model~\cite{wang2025pi3}, all images are resized to $224 \times 224$. For AnySplat, by contrast, we keep the training resolution at $448 \times 448$.
% (i) high-diversity, real-world datasets dominate the mix, so that the bulk of training comes from the real distribution;
% and (ii) synthetic datasets supplement the mix with outdoor, city-scale, and wide-baseline regimes that are under-represented in available real captures. The input resolution and view-count sampling follow the pretraining regime of each host backbone. 
% For Pi3X and
% Distill Pi3X, following the training setup of the original Pi3X model~\cite{pi3}, all images are resized to $224 \times 224$, and the number of input views per sample is drawn from $[2, 24]$ to expose the model to a wide range of capture densities. 
% For AnySplat, by contrast, we keep the training resolution at $448 \times 448$ and draw the number of input views from $[2, 8]$.

\paragraph{Baselines.}
We position MetaGrad as a \emph{plug-in training strategy} that applies to any feed-forward Gaussian-prediction model equipped with a differentiable renderer. 
To demonstrate its generality, we attach it to three representative backbones.
% , and evaluate each backbone under two matched-budget settings: \emph{Vanilla}, in which the Gaussian head is fine-tuned for the same $2{,}000$ outer-loop steps using only the original supervised objective, and \emph{w/ MetaGrad}, in which the same head is fine-tuned for the same number of steps with our meta-gradient objective. 
% This controlled comparison ensures that any observed gain is attributable to the MetaGrad signal rather than to the additional fine-tuning budget alone. 
The three backbones are:
\begin{enumerate}
    \item \textbf{AnySplat}~\cite{jiang2025anysplat}, a state-of-the-art feed-forward 3DGS model that jointly predicts Gaussians and camera parameters in a single forward pass.

    \item \textbf{Pi3X}~\cite{wang2025pi3}, one of the most accurate feed-forward geometry foundation models to date. Since Pi3X does not natively predict Gaussians, we attach a lightweight DPT-style~\cite{ranftl2021dpt} Gaussian head on top of its predicted point cloud so that it produces renderable 3D Gaussians; the head architecture and pre-training protocol are detailed in the \textit{Appendix}.

    \item \textbf{Distill Pi3X}, a distilled variant of Pi3X.
    Its parameter count is roughly 45\% that of Pi3X. 
    % with a corresponding drop in camera-pose and point-cloud accuracy; 
    Full architecture and distillation details are given in the \textit{Appendix}.
\end{enumerate}

In addition, we include \textbf{InstantSplat}~\cite{InstantSplat} as a traditional predict-then-refine baseline in the quantitative summary. 
Since it is not a MetaGrad-compatible Gaussian-head backbone in our setup, Tab.~\ref{tab:postopt_quantitative} reports only its vanilla results.

\paragraph{Training protocol.}
Starting from the pretrained checkpoint of each backbone, we fine‑tune only the Gaussian prediction head for $2{,}000$ iterations while keeping the host backbone entirely frozen.
For AnySplat, we optimize its native Gaussian head; for Pi3X and Distill Pi3X, we optimize only the appended DPT‑style Gaussian head described above.
In the MetaGrad variant, each outer‑loop iteration unrolls a detached inner refinement trajectory whose length is randomly sampled from $[50, 500]$ steps; anchor states are recorded every $40$ steps along that trajectory.

% For each MetaGrad-compatible backbone, we initialize from its pretrained checkpoint and carry out a $2{,}000$-step additional fine-tuning stage in which only the Gaussian head is updated.
% The host backbone is kept frozen throughout: for AnySplat, we optimize its Gaussian-prediction head, whereas for Pi3X and Distill Pi3X, we optimize only the attached DPT-style Gaussian head introduced above.
% For the MetaGrad variant, each outer-loop step rolls out a detached inner post-optimization trajectory with a randomly sampled horizon between $50$ and $500$ steps, and anchor states are sampled every $40$ steps along the trajectory.

\paragraph{Training cost.}
Tab.~\ref{tab:training_cost} reports the computational cost of the different training stages used in our experiments. Baseline training, which may involve full backbone optimization, requires substantially more GPU‑hours. In contrast, the MetaGrad fine‑tuning stage requires only a single GPU and completes in a few GPU‑hours for all backbones.

% Tab.~\ref{tab:training_cost} reports the computational cost of the additional fine‑tuning stage used in our main experiments, which requires only a few GPU‑hours in all cases.

% All wall-clock times are measured on a single NVIDIA RTX~5090 GPU with $32$\,GB of memory, and all post-optimization benchmarks reported in Sec.~\ref{section:exp:main} and Sec.~\ref{section:exp:ablation} are run on the same GPU class.
% Tab.~\ref{tab:training_cost} shows that this additional fine-tuning stage requires only a few GPU-hours in all main experiments. 

\begin{table*}[t]
    \centering
    \small
    \caption{
     Computational cost of the different training stages used in our experiments.
    }
    \setlength{\tabcolsep}{4pt}
    \begin{threeparttable}
    \begin{tabular*}{\linewidth}{@{\extracolsep{\fill}} l l c c c c @{}}
        \toprule
        Method & Backbone & Trainable part & GPUs & Wall-clock & GPU hours \\
        \midrule
        \multirow{3}{*}{\textbf{w/o MetaGrad}} 
            & AnySplat$^1$    & Backbone, GS head & $16\times$ NVIDIA~A800 & 48.00 & 768.00 \\
            & Pi3X            & GS head            & $2\times$ RTX~5090    & 23.11  & 46.22    \\
            & Distill Pi3X    & Backbone, GS head & $6\times$ NVIDIA~H20   & 124.31 & 745.86 \\
        \midrule
        \multirow{3}{*}{\textbf{MetaGrad Finetune}} 
            & AnySplat        & GS head            & $1\times$ RTX~5090     & 6.86  & 6.86  \\
            & Pi3X            & GS head            & $1\times$ RTX~5090     & 6.11  & 6.11  \\
            & Distill Pi3X    & GS head            & $1\times$ RTX~5090     & 4.41  & 4.41  \\
        \bottomrule
    \end{tabular*}
    \begin{tablenotes}
        \footnotesize
        \item[1] Results from original paper.
    \end{tablenotes}
    \end{threeparttable}
    \label{tab:training_cost}
    % \Description{table training cost}
\end{table*}

\paragraph{Evaluation benchmarks.}
We evaluate MetaGrad on five datasets:
CO3Dv2~\cite{reizenstein2021co3d},
DL3DV-10K~\cite{ling2024dl3dv},
GTA-SfM~\cite{wang2020gtasfm},
ScanNet++~\cite{yeshwanth2023scannetpp},
and TartanAir~\cite{wang2020tartanair}.
For evaluation, a total of $50$ scenes are sampled from these datasets, with any scene appearing in the training split explicitly excluded to keep the evaluation set strictly scene-disjoint.
Each scene contains $16$ images, from which every fourth image is held out as a test view and the remaining $12$ images are treated as training views. 

Following AnySplat~\cite{jiang2025anysplat} and InstantSplat~\cite{InstantSplat}, we report results under a \emph{pose-aligned AnySplat-style evaluation protocol}. Each scene is first passed once through the feed-forward model using all $16$ images to obtain a globally consistent scene prediction and camera initialization. Post-optimization then retains only the Gaussians associated with the $12$ training views: the test-view Gaussians are discarded, and the test-view RGB images are never used as photometric supervision for updating the Gaussian parameters.

% During post‑optimization, we enable camera optimization for the training views, as is common in other refinement based 3DGS pipelines~\cite{InstantSplat}.
% This can cause the refined training-view cameras to drift away from the original feed-forward estimates, in which case errors on held-out test views may become dominated by camera misalignment rather than by the underlying scene representation. Fig.~\ref{fig:pose_refine} illustrates this effect: with the same optimized Gaussian scene, an unrefined test pose can create structured residuals that mainly reflect camera misalignment rather than intrinsic Novel View Synthesis (NVS) quality. We therefore refine only the held-out test-camera parameters against the frozen Gaussian scene before computing NVS metrics. This evaluation-time pose refinement does not modify the Gaussian parameters and does not alter the training-view refinement trajectory. Unless otherwise stated, all NVS numbers in this paper are reported under this protocol.

All methods are evaluated on exactly the same $50$ scenes, the same train/test view splits, and the same evaluation-time pose-alignment procedure to ensure a fair comparison.

% \begin{figure}[t]
%     \centering
%     \includegraphics[width=1.0\linewidth]{figure/4_exp/poserefine.pdf}
%     \caption{Necessity of evaluation‑time pose refinement.
%     Both renderings originate from the same optimized Gaussian scene and differ only in the test‑view camera.
%     Without the proposed per‑test‑view pose refinement, the lower PSNR primarily reflects residual misalignment between the held‑out test camera and the optimized scene, visible as structured residuals in the difference image, rather than a decline in scene representation quality.
%     }
%     \label{fig:pose_refine}
%     \Description{pose refine.}
% \end{figure}

\paragraph{Post-optimization protocol.}
During post-optimization, structural operations such as densification, cloning, splitting, and pruning are disabled; only the attributes of the existing Gaussians are updated.
The inner post-optimization optimizer, loss terms, learning-rate schedule, and per-parameter learning rates are kept fixed throughout all the experiments.

\subsection{MetaGrad Gains}
\label{section:exp:main}

\begin{table*}[t]
    % \begin{threeparttable}
    \centering
    \scriptsize
    \caption{Quantitative post-optimization results.
    PSNR, SSIM, and LPIPS are reported on the same held-out evaluation views over $0$--$2{,}000$ refinement steps.
    For each MetaGrad-trained model, \emph{$\approx$ steps} denotes the number of refinement steps required to match the PSNR of its vanilla counterpart at each budget.
    % while \emph{Scenes better} reports the percentage of scenes where the MetaGrad-trained model achieves higher PSNR.
    Following the evaluation protocol of the original backbones, AnySplat is evaluated at $448{\times}448$ resolution, while Pi3X and Distill Pi3X are evaluated at $224{\times}224$ resolution.
    }
    \setlength{\tabcolsep}{2.5pt}
    \resizebox{\textwidth}{!}{%
    \begin{tabular}{@{}l*{15}{c}*{3}{c}*{3}{c}@{}}
        \toprule
        \multirow{2}{*}{\textbf{Method}}
        & \multicolumn{5}{c}{\textbf{PSNR} $\uparrow$}
        & \multicolumn{5}{c}{\textbf{SSIM} $\uparrow$}
        & \multicolumn{5}{c}{\textbf{LPIPS} $\downarrow$}
        & \multicolumn{3}{c}{\textbf{$\approx$ steps} $\downarrow$} \\
        \cmidrule(lr){2-6}\cmidrule(lr){7-11}\cmidrule(lr){12-16}\cmidrule(lr){17-19}\cmidrule(l){20-22}
        & 0 & 200 & 500 & 1k & 2k
        & 0 & 200 & 500 & 1k & 2k
        & 0 & 200 & 500 & 1k & 2k
        & 500 & 1k & 2k\\
        
        \midrule
        
        {\textbf{InstantSplat}}
        & 18.77 & 23.55 & 25.20 & 25.63 & 25.91
        & 0.640 & 0.732 & 0.771 & 0.779 & 0.771
        & 0.389 & 0.236 & 0.202 & 0.193 & 0.194
        & - & - & -\\
        \multicolumn{22}{@{}l}{\textbf{AnySplat}} \\
        \quad Vanilla
        & \best{21.66} & 26.17 & 26.88 & 27.02 & 27.03
        & \best{0.736} & 0.838 & 0.844 & 0.839 & 0.833
        & \best{0.254} & \best{0.166} & 0.153 & 0.156 & 0.163
        & - & - & - \\
        \quad + MetaGrad
        & 20.70 & \best{26.36} & \best{27.32} & \best{27.60} & \best{27.68}
        & 0.698 & \best{0.839} & \best{0.852} & \best{0.850} & \best{0.846}
        & 0.299 & 0.172 & \best{0.148} & \best{0.144} & \best{0.146}
        & 300 & 350 & 350\\

        \midrule
        
        \multicolumn{22}{@{}l}{\textbf{Pi3X}} \\
        \quad Vanilla
        & \best{21.29} & 23.88 & 24.48 & 24.86 & 24.97
        & \best{0.675} & \best{0.775} & 0.782 & 0.784 & 0.777
        & \best{0.231} & \best{0.168} & \best{0.164} & 0.165 & 0.171
        & - & - & -\\
        \quad + MetaGrad
        & 18.84 & \best{23.92} & \best{24.92} & \best{25.44} & \best{25.81}
        & 0.562 & 0.762 & \best{0.783} & \best{0.788} & \best{0.792}
        & 0.365 & 0.193 & 0.167 & \best{0.158} & \best{0.153}
        & 300 & 500 & 500\\
        \multicolumn{22}{@{}l}{\textbf{Distill Pi3X}} \\
        \quad Vanilla
        & \best{20.30} & 22.89 & 23.28 & 23.50 & 23.61
        & \best{0.597} & 0.717 & 0.725 & 0.729 & 0.727
        & \best{0.287} & \best{0.175} & 0.173 & 0.175 & 0.180
        & - & - & -\\
        \quad + MetaGrad
        & 19.44 & \best{23.02} & \best{23.61} & \best{23.86} & \best{24.05}
        & 0.580 & \best{0.721} & \best{0.737} & \best{0.741} & \best{0.743}
        & 0.326 & 0.181 & \best{0.170} & \best{0.168} & \best{0.168}
        & 300 & 400 & 500\\
        
        \bottomrule
    \end{tabular}%
    }
    % \begin{tablenotes}
    % \footnotesize
    %     \item[1] xx.
    % \end{tablenotes}
    % \end{threeparttable}
    \label{tab:postopt_quantitative}
    % \Description{main result table}
\end{table*}

\begin{figure}
    \centering
    \includegraphics[width=1.0\linewidth]{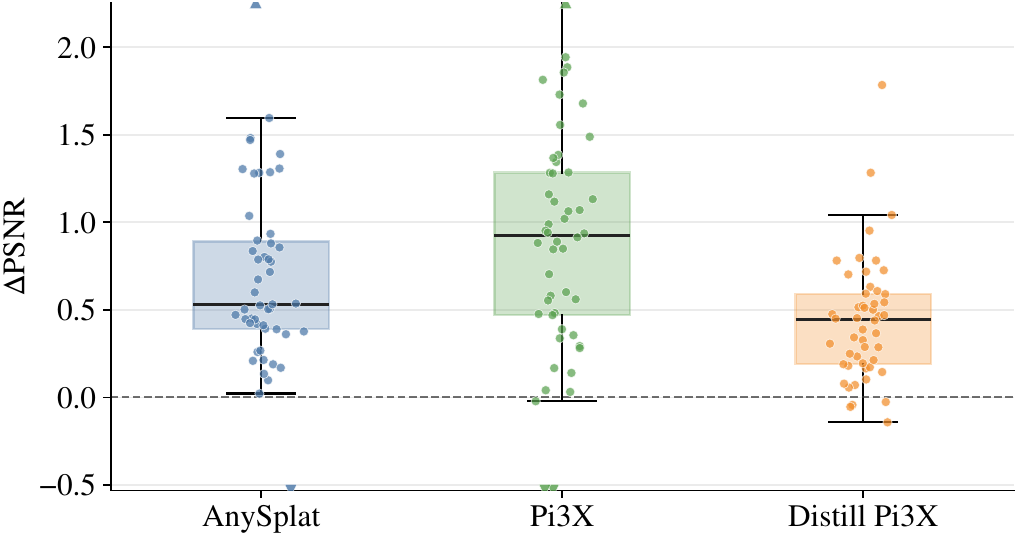}
    \caption{Per-scene PSNR gain at $2{,}000$ post-optimization steps.
    For each backbone, each point denotes one evaluation scene and the vertical axis reports $\mathrm{PSNR}_{\text{MetaGrad}} - \mathrm{PSNR}_{\text{vanilla}}$ at $2{,}000$ post-optimization steps. 
    Box plots summarize the scene-level distribution, and the
    inset statistics report the mean, median, and positive-scene ratio.}
    \label{fig:pre_scene_delta_distribution_psnr_2000}
    % \Description{per scene psnr delta}
\end{figure}

We compare checkpoints trained with and without MetaGrad.
Following the protocol of Sec.~\ref{section:exp:setup}, we measure the evolution of PSNR, SSIM, and LPIPS on held‑out test views over a $2{,}000$‑step post‑optimization window.
Quantitative summaries are reported in Tab.~\ref{tab:postopt_quantitative}. 
% Full training trajectories are presented in \textit{Appendix}.
% and the full training trajectories are shown in Fig.~\ref{fig:main_results}.

% For each of the three feed-forward backbones---AnySplat, Pi3X, and Distill Pi3X---we fine-tune only the Gaussian head for $2{,}000$ outer-loop steps. 
% We then compare the checkpoint obtained with MetaGrad against a \emph{vanilla} checkpoint obtained by fine-tuning the same head for the same number of steps without MetaGrad.
% Following the protocol of Sec.~\ref{section:exp:setup}, we measure the evolution of PSNR, SSIM, and LPIPS on held-out test views over a $2{,}000$-step post-optimization window. 
% Quantitative summaries are reported in Tab.~\ref{tab:postopt_quantitative}, and full trajectories are reported in Fig.~\ref{fig:main_results}.

\paragraph{Post-optimization quality.}
Under the same post‑optimization budgets, MetaGrad consistently outperforms vanilla across all three backbones and all three metrics.
% (Fig.~\ref{fig:main_results}, right insets).
The benefit of MetaGrad is most pronounced on Pi3X, where PSNR increases by nearly $0.9$\,dB ($24.97$ vs. $25.81$), along with consistent gains in SSIM and LPIPS. The advantage also holds on AnySplat and Distill Pi3X.
Despite substantial differences among the three backbones in feed‑forward architecture, parameter count, and training data, the direction of improvement remains identical.

% Under matched post-optimization budgets, MetaGrad consistently outperforms the vanilla baseline across all three backbones and all three metrics (Fig.~\ref{fig:main_results}, right insets). 
% The gain is largest on Pi3X, where PSNR improves by nearly $0.9$\,dB ($24.97$ vs. $25.81$), with corresponding improvements in SSIM and LPIPS; the advantage also holds on AnySplat and Distill Pi3X.
% Despite substantial differences among the three backbones in feed‑forward architecture, parameter count, and training data, the direction of improvement remains identical.

This gain is not driven by a small number of favorable scenes. Fig.~\ref{fig:pre_scene_delta_distribution_psnr_2000} shows that, for all three backbones, the per-scene PSNR differences relative to vanilla are broadly shifted to the positive values, while the inset mean, median, and positive-scene ratio all remain favorable to MetaGrad. 
This scene-level picture is consistent with the ``Scenes better'' columns in Tab.~\ref{tab:postopt_quantitative}, where MetaGrad yields higher PSNR on the large majority of evaluation scenes under the same post-optimization step budget.
Together, these statistics confirm that the average gain reflects broad scene‑level improvement rather than a handful of outliers within this 50‑scene paired benchmark.

\paragraph{Convergence speed.}
The per‑step metrics reported in Tab.~\ref{tab:postopt_quantitative} show that MetaGrad‑trained initializations converge markedly faster than their vanilla counterparts.
Although MetaGrad may start from a slightly lower immediate rendering quality, its post‑optimization trajectory rises with a consistently steeper slope, overtaking the vanilla curve well before the final budget and continuing to a higher end‑of‑budget plateau.
This accelerated convergence follows directly from the MetaGrad objective: the Gaussian head is trained to produce initializations that the downstream optimizer can improve rapidly, rather than to match the target views at step zero.
The resulting predictions are thus \emph{post‑optimization‑friendly}: they occupy a region of the loss landscape from which per‑step refinement is more effective, yielding both faster progress and a higher final quality under the same optimization budget. To provide further intuitions, the full training trajectories are presented in \textit{Appendix}.

\paragraph{End-to-End reconstruction comparison.}

Tab.~\ref{tab:time_psnr} separates the fixed feed-forward cost from the additional post-optimization budget: each backbone header reports the shared inference time ($0.72$\,s for AnySplat, $0.30$\,s for Pi3X, and $0.23$\,s for Distill Pi3X), the \emph{inference} column reports the zero-step feed-forward prediction, and $+1$\,s to $+20$\,s denote additional post-optimization time after inference.
Under a $+1$\,s post-optimization budget, the two variants are already comparable; once the extra budget reaches $+2$\,s, MetaGrad outperforms vanilla on all three backbones, and the gap continues to widen as more post-optimization time becomes available.
At $+20$\,s of post-optimization, for example, PSNR improves from $27.02$ to $27.61$ on AnySplat, from $24.97$ to $25.81$ on Pi3X, and from $23.61$ to $24.05$ on Distill Pi3X, showing that ForeSplat yields a better practical reconstruction quality-time trade-off.

\begin{table}[t]
    \centering
    \caption{
        PSNR performance under varying post-optimization wall-clock time budgets. 
        The number following each backbone name denotes its feed-forward inference time, shared by both Vanilla and MetaGrad. 
        The \emph{inference} column reports the zero-step feed-forward prediction, while $+1$\,s to $+20$\,s denote additional post-optimization time allocated after inference.
    }
    \setlength{\tabcolsep}{2pt}
    \begin{tabular*}{\linewidth}{@{\extracolsep{\fill}} lcccccc @{}}
        \toprule
        \multirow{2}{*}{\textbf{Method}}
         & \multicolumn{6}{c}{\textbf{PSNR}~[dB] $\uparrow$}\\
        \cmidrule(lr){2-7}
        & inference & +1s & +2s & +5s & +10s & +20s \\
        \midrule
        \multicolumn{7}{@{}l}{\textbf{AnySplat} (inference: 0.72s)} \\
        \quad Vanilla & 21.66 & 24.85 & 25.56 & 26.56 & 26.97 & 27.02 \\
        \quad + MetaGrad & 20.70 & 24.98 & 25.81 & 26.92 & 27.43 & 27.61 \\
        \midrule
        \multicolumn{7}{@{}l}{\textbf{Pi3X} (inference: 0.30s)} \\
        \quad Vanilla & 21.29 & 23.48 & 23.88 & 24.47 & 24.87 & 24.97 \\
        \quad + MetaGrad & 18.84 & 23.28 & 23.92 & 24.92 & 25.55 & 25.81 \\
        \multicolumn{7}{@{}l}{\textbf{Distill Pi3X} (inference: 0.23s)} \\
        \quad Vanilla & 20.30 & 22.36 & 22.89 & 23.29 & 23.50 & 23.61 \\
        \quad + MetaGrad & 19.44 & 22.38 & 23.02 & 23.60 & 23.87 & 24.05 \\
        \bottomrule
    \end{tabular*}
    \label{tab:time_psnr}
    % \Description{table: time-psnr}
\end{table}

\paragraph{Qualitative Comparison}
\begin{figure*}[p]
    \centering
    \includegraphics[width=1.0\linewidth]{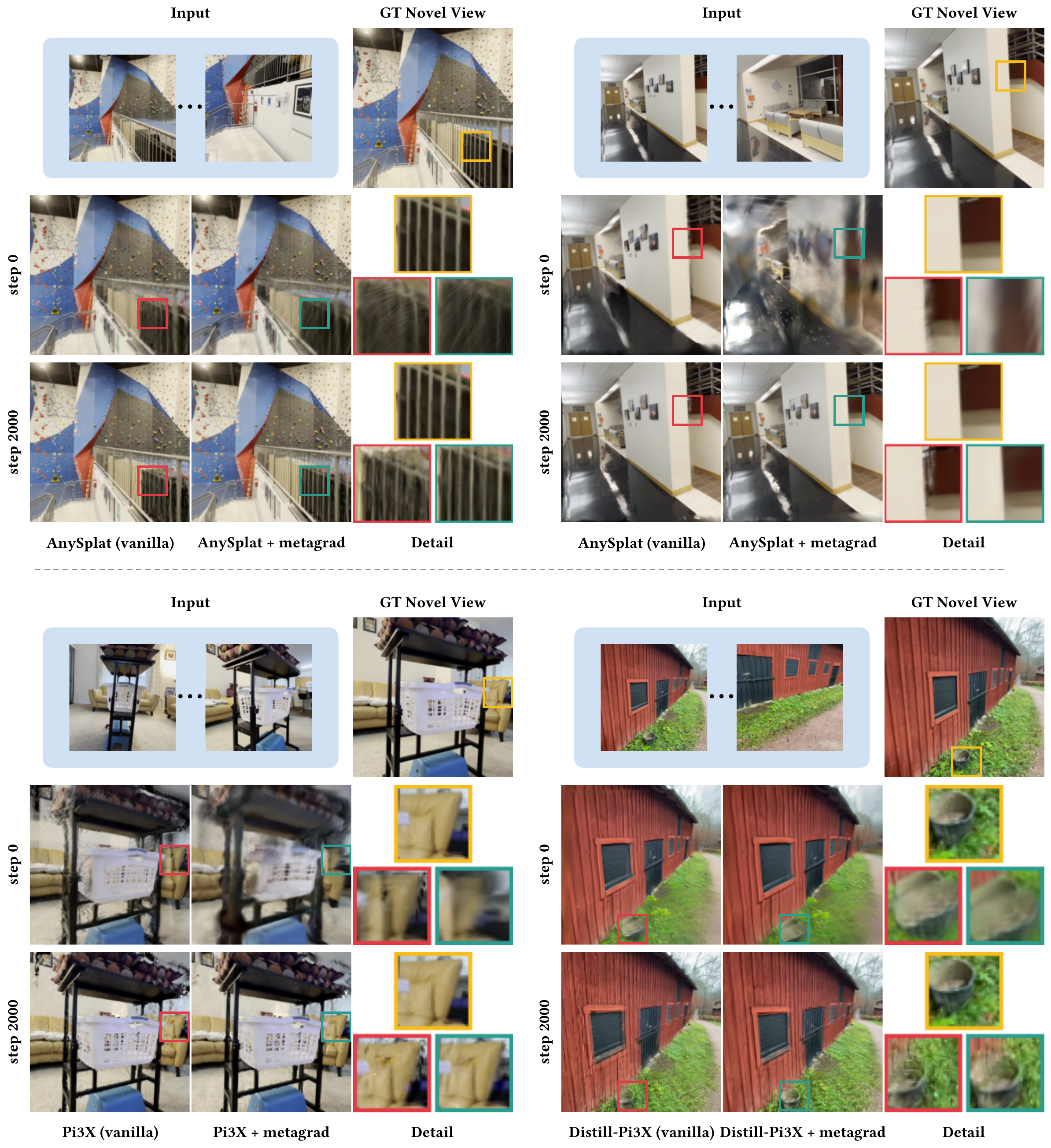}
    \caption{Qualitative comparison of vanilla and MetaGrad before and after post-optimization.
    MetaGrad starts from weaker zero‑step renderings, yet after post‑optimization, it consistently attains cleaner structure and sharper appearance than vanilla, demonstrating a more post‑optimization‑friendly initialization.
    }
    % \Description{Qualitative comparison.}
    \label{fig:compare}
\end{figure*}

\begin{figure*}[p]
    \centering
    \includegraphics[width=0.95\linewidth]{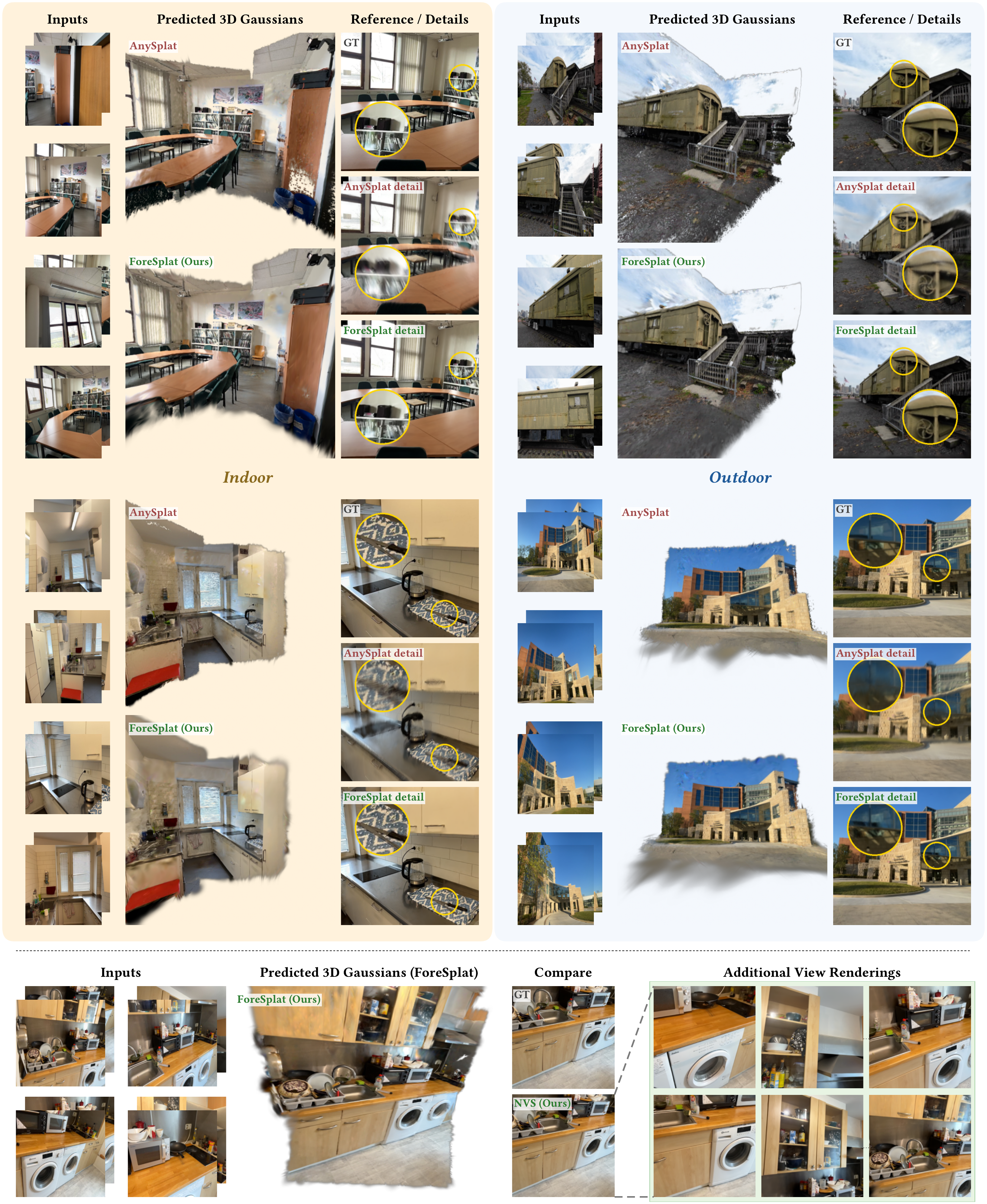}
    \caption{Qualitative comparison of 3D reconstruction quality between AnySplat and our ForeSplat (ours). The upper block shows two indoor (left) and two outdoor (right) scene. The bottom row showcases a single scene rendered from multiple novel viewpoints by our method.}
    % \Description{Qualitative comparison2.}
    \label{fig:compare2}
\end{figure*}

As shown in Fig.~\ref{fig:compare} and Fig.~\ref{fig:compare2}, the proposed ForeSplat produces optimization‑friendly initializations that refine into higher‑fidelity scene representations within seconds, and it generalizes these gains across diverse scenes and backbones.
Fig.~\ref{fig:compare} demonstrates this optimization‑friendly behavior explicitly: even though ForeSplat may start from slightly weaker zero‑step renderings, after post‑optimization it consistently recovers finer details and sharper structures than vanilla, maintaining this advantage even on the lightweight distilled backbone, which illustrates end‑to‑end high‑quality 3D reconstruction in a compact setting.
Fig.~\ref{fig:compare2} further compares our method with the AnySplat baseline~\cite{jiang2025anysplat}; across a wide range of indoor and outdoor scenes, ForeSplat delivers higher reconstruction quality, yielding cleaner geometry and more faithful appearance under the same evaluation protocol.

% The post-optimization trajectories in Fig.~\ref{fig:main_results} exhibit a clear two-stage pattern. 
% \emph{Early} in post-optimization (left inset), MetaGrad yields lower PSNR/SSIM and higher LPIPS than vanilla; 
% the two trajectories then cross, and from the \emph{middle} of post-optimization onward the reported MetaGrad curves remain above vanilla, ending with a clear lead at the end of the plotted budget (right inset). 
% This pattern is structurally consistent with the MetaGrad objective: the loss does not reward instantaneous pixel-wise agreement between the head's output and the target views, but rather the final reconstruction quality reachable from that output after a fixed number of post-optimization steps. 
% As a result, the trained Gaussian head produces initial Gaussians that score lower under supervised rendering metrics at the start, but converge to a higher plateau under post-optimization. We refer to this property as a \emph{post-optimization-friendly initialization}.

\subsection{Ablation Studies}
\label{section:exp:ablation}

In this section, we ablate two central design choices of MetaGrad: the loss-balancing coefficient $\lambda$ in Eq.~\ref{eq:total_loss} and the anchor stride $\Delta$ in Eq.~\ref{eq:anchor_set}. 
The $\lambda$ sweep, repeated on all three backbones, tests whether the gains and trajectory-shape changes reported in Sec.~\ref{section:exp:main} are genuinely driven by the meta-gradient signal rather than by the additional fine-tuning compute, and whether this trend is stable across architectures. 
We then study the anchor-sampling design on AnySplat, comparing different $\Delta$ values and a Reptile-style first-order alternative to assess how the density of anchor supervision affects post-optimization quality.

\subsubsection{Effect of the Meta-Gradient Weight $\lambda$}

\paragraph{Sweep configuration.}

We sweep $\lambda \in \{0.0, 0.25, 0.5, 0.75, 1.0\}$, with each value corresponding to an independent $2{,}000$-step outer-loop run. 
The two endpoints carry distinct interpretations. At $\lambda = 1.0$, the meta-gradient term is disabled entirely, so the procedure reduces to purely supervised fine-tuning for $2{,}000$ additional steps under the original supervised loss. This is exactly the \emph{vanilla} setting used in \ref{section:exp:main}. At $\lambda = 0.0$, the training uses the meta-gradient term exclusively (\emph{pure-meta}). 
For every variant we record the full PSNR/SSIM/LPIPS trajectories over the same $2{,}000$-step post-optimization window. 
% Fig.~\ref{fig:lambda_sweep}
The full post-optimization trajectories together with the 2,000-step summaries across all three backbones are provided in the Appendix, while 
Tab.~\ref{tab:ablation lambda} provides the detailed Pi3X PSNR values at post-optimization steps $0$, $200$, $500$, $1{,}000$, and $2{,}000$.

\begin{table}[t]
    \centering
    \caption{
        Ablation of $\lambda$ on the Pi3X backbone. Colored PSNR cells mark the top three values in each step column, indicated by \protect\colorbox{color_1st}{1st}, \protect\colorbox{color_2nd}{2nd}, and \protect\colorbox{color_3rd}{3rd}, respectively.
    }
    \setlength{\tabcolsep}{4pt}
    \begin{tabular*}{\linewidth}{@{\extracolsep{\fill}} lccccc @{}}
        \toprule
        \multirow{2}{*}{\textbf{Method}}
         & \multicolumn{5}{c}{\textbf{PSNR} $\uparrow$}\\
        \cmidrule(lr){2-6}
        & 0 & 200 & 500 & 1k & 2k \\
        \midrule
        baseline       & \2rk{21.30} & 23.71 & 24.27 & 24.62 & 24.82 \\
        $\lambda=1.00$ & \3rk{21.29} & 23.88 & 24.48 & 24.86 & 24.97 \\
        $\lambda=0.75$ & \1rk{21.31} & 23.94 & 24.62 & 24.96 & 25.07 \\
        $\lambda=0.50$ & 21.23 & \2rk{24.02} & \3rk{24.71} & \3rk{25.07} & \3rk{25.28} \\
        $\lambda=0.25$ & 20.99 & \1rk{24.03} & \2rk{24.79} & \2rk{25.19} & \2rk{25.41} \\
        $\lambda=0.00$ & 18.84 & \3rk{23.92} & \1rk{24.92} & \1rk{25.44} & \1rk{25.81} \\
        \bottomrule
    \end{tabular*}
    \label{tab:ablation lambda}
    % \Description{table: ablation lambda}
\end{table}

\paragraph{Dependence on the meta-gradient weight.}

A consistent trend emerges across all three backbones: 
% as the meta-gradient weight $1 - \lambda$ increases, end-of-budget PSNR and SSIM rise monotonically and LPIPS falls monotonically (Fig.~\ref{fig:lambda_sweep}). 
relative to the purely supervised endpoint $\lambda = 1.0$, increasing the meta-gradient weight improves end-of-budget PSNR and SSIM and lowers LPIPS, with the strongest results appearing in the strong-meta region of the sweep.
% (Fig.~\ref{fig:lambda_sweep}). 
The full sweep from $\lambda{=}1$ to $\lambda{=}0$ moves PSNR by roughly $0.4$--$0.8$\,dB depending on the backbone, 
% with the curves remaining smooth and directionally consistent throughout. 
with the tabulated checkpoints remaining monotonic and directionally consistent

\paragraph{Effect on trajectory shape.}

% The monotonicity of the end-of-budget metrics is mirrored in the shape of the post-optimization trajectory. 
% As $1 - \lambda$ grows, the early post-optimization deficit deepens, and the late-stage plateau lifts further.
% This monotonic dependence shows that the two-stage ``below vanilla early, above vanilla late'' pattern described in Sec.~\ref{section:exp:main} is not a binary phenomenon but a controllable property whose strength grows continuously with the weight of the meta-gradient term.
% ---------------------
The same end-of-budget trend is mirrored in the shape of the post-optimization trajectory. 
As $1 - \lambda$ grows, the early post-optimization deficit deepens, while the late-stage plateau generally lifts.
This dependence shows that the two-stage ``below vanilla early, above vanilla late'' pattern described in Sec.~\ref{section:exp:main} is not a binary phenomenon but a controllable property whose strength varies with the weight of the meta-gradient term.

\subsubsection{Effect of the Anchor Sampling Stride $\Delta$}

\begin{table}[t]
    \centering
    \caption{
    Ablation of the anchor stride $\Delta$ in MetaGrad on the AnySplat backbone.
    We report PSNR at post-optimization steps $0$, $200$, $500$, $1{,}000$, and $2{,}000$. In Eq.~\ref{eq:anchor_set}, $\Delta$ denotes the anchor-sampling interval along the inner optimization trajectory.
    \emph{Vanilla} denotes supervised fine-tuning without MetaGrad, \emph{Reptile} denotes the Reptile-style first-order baseline defined in \cref{eq:reptile_theta}, and \emph{Baseline} denotes the original model without fine-tuning. 
    \emph{GPU Hours} denotes the time required by each fine-tuning-based variant to complete the same 2,000-step fine-tuning run on a single RTX 5090.
    Colored PSNR cells use the same ranking convention as in Tab.~\ref{tab:ablation lambda}.}
    \setlength{\tabcolsep}{3pt}
    \begin{tabular*}{\linewidth}{@{\extracolsep{\fill}} lcccccc @{}}
        \toprule
        \multirow{2}{*}{\textbf{Method}}
        & \multicolumn{5}{c}{\textbf{PSNR} $\uparrow$} & \multirow{2}{*}{\textbf{GPU Hours}} \\
        \cmidrule(lr){2-6}
        & 0 & 200 & 500 & 1k & 2k \\
        \midrule
        Baseline & \2rk{21.90} & \3rk{26.16} & \3rk{26.91} & 27.03 & 27.05 & -\\
        Vanilla & \3rk{21.66} & \2rk{26.17} & 26.88 & 27.02 & 27.03 & 0.98 \\
        Reptile & \1rk{22.53} & \3rk{26.16} & 26.88 & 26.97 & 27.00 & 8.05 \\
        $\Delta$1 & 21.32 & 26.11 & 26.86 & 27.13 & 27.20 &  22.37 \\
        $\Delta$20 & 21.09 & 26.15 & \2rk{26.92} & \2rk{27.19} & \2rk{27.28} & 6.54 \\
        $\Delta$40 & 20.70 & \1rk{26.36} & \1rk{27.32} & \1rk{27.60} & \1rk{27.68} & 6.86 \\
        $\Delta$80 & 21.20 & 26.11 & \3rk{26.91} & \3rk{27.17} & \3rk{27.19} & 6.75 \\
        \bottomrule
        
    \end{tabular*}
    
    \label{tab:metagrad_variants}
    % \Description{table: metagrad_variants}
\end{table}

The hyperparameter $\Delta$ in Eq.~\ref{eq:anchor_set} controls how sparsely anchor states are sampled along the inner optimization trajectory. To examine its practical effect, we keep all other training settings fixed and sweep over $\Delta$, while also comparing against the vanilla baseline and a Reptile-style first-order alternative (Eq.~\ref{eq:reptile_theta}), which is equivalent to simply summing the gradients over all inner-loop steps.
A denser interval ($\Delta = 1$) overemphasizes highly correlated neighboring inner states, whereas a sparser interval ($\Delta = 80$) loses coverage of the optimization trajectory; $\Delta = 40$ balances these two effects.
The Reptile-style alternative attains a stronger step‑0 initialization, but fails to translate that early advantage into better reconstruction quality at larger optimization budgets. This aligns with our claim that MetaGrad should optimize the post-optimization trajectory rather than the immediate feed‑forward prediction alone.

% In Eq.~\ref{eq:anchor_set}, $\Delta$ denotes the sampling interval used to select anchor states along the detached inner optimization trajectory. To test whether this design choice matters in practice, we fix the rest of the training setup and vary $\Delta$, while also comparing against the vanilla baseline and a Reptile-style first-order alternative (Eq.~\ref{eq:reptile_theta}) which is equivalent to sum up gradients from all steps. 
% Tab.~\ref{tab:metagrad_variants} shows that the default choice $\Delta = 40$ consistently gives the highest PSNR at every reported post-optimization budget. 
% Compared with denser anchoring ($\Delta = 1$), it avoids over-emphasizing highly correlated neighboring inner states; compared with sparser anchoring ($\Delta = 80$), it preserves stronger coverage of the optimization trajectory. 

% The Reptile-style alternative attains a stronger step-$0$ initialization, but fails to translate that advantage into better late-budget reconstruction quality, which is consistent with our claim that MetaGrad should optimize the post-optimization trajectory rather than the immediate feed-forward prediction alone.
%% sec/5_conclusion.tex
\section{Application: Toward Edge 3D Reconstruction Cameras}

\begin{figure}[H]
    \centering
    \includegraphics[width=1.0\linewidth]{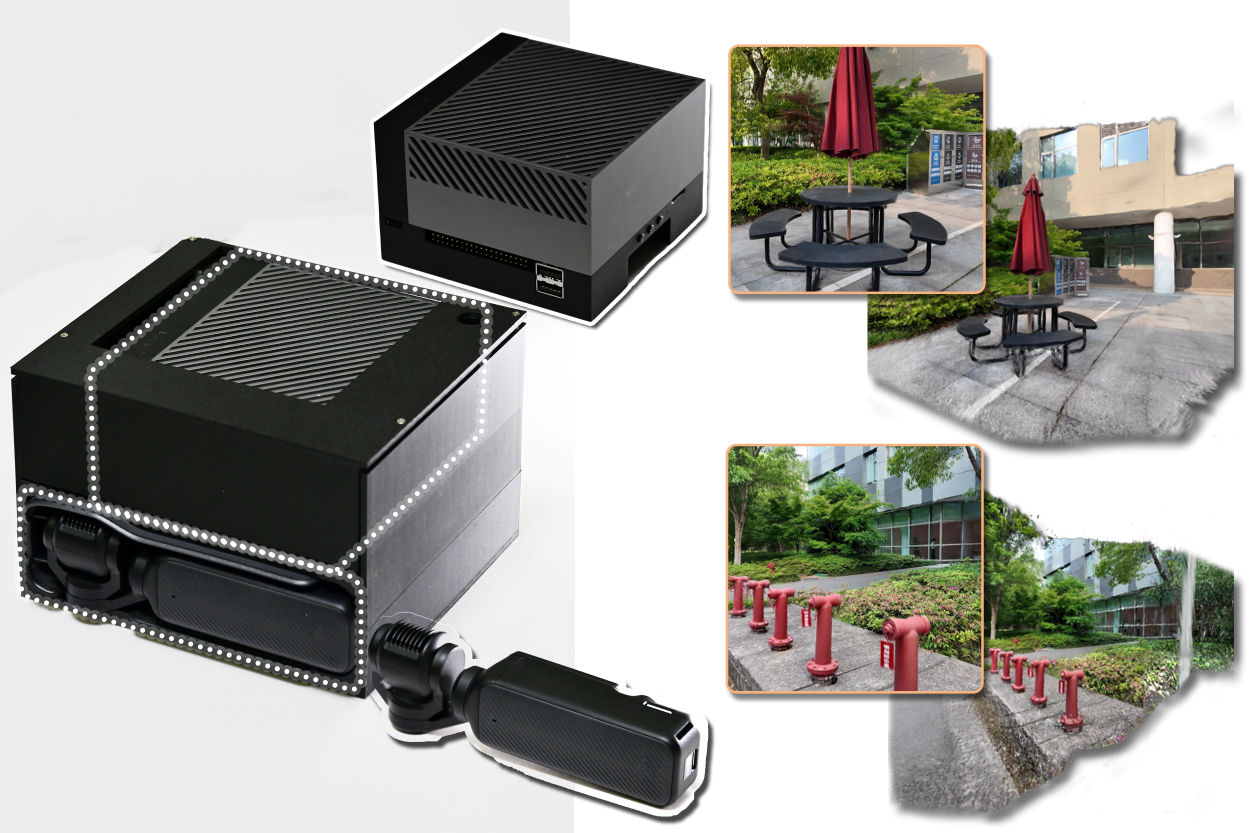}
    \caption{A lightweight capture‑and‑refine camera that produces renderable 3D scene representations on device within seconds.}
    \label{fig:placeholder}
\end{figure}

Beyond improving post-optimization metrics, 
ForeSplat points to a practical deployment scenario: lightweight, high‑quality, end‑to‑end 3D reconstruction within seconds in a compact capture‑and‑refine pipeline.
Existing feed-forward 3DGS models offer fast scene prediction, but achieving high reconstruction quality often requires either a large backbone or a costly per-scene optimization process. ForeSplat changes this trade-off by shifting part of the reconstruction burden from the feed-forward model itself to a short, optimization-aware refinement stage. Instead of requiring the network to directly predict a final high-fidelity radiance field in a single forward pass, the model is trained to produce an initialization that can be rapidly improved by downstream differentiable rendering-based optimization.

This property is particularly important for edge-side 3D reconstruction. Since the post-optimization stage can compensate for part of the representational limitations of the feed-forward predictor, ForeSplat makes it feasible to use a compact distilled backbone rather than a full-scale foundation model. The distilled model provides an efficient initial reconstruction under limited memory and compute, while the subsequent lightweight refinement recovers missing details and improves rendering fidelity. In this way, ForeSplat relaxes the dependency on large server-side models and opens the possibility of running practical 3D reconstruction on edge AI hardware.

A representative application is a 3D reconstruction camera. Such a device would capture a short burst of images, infer an initial 3D Gaussian representation directly on-device, and then perform a small number of refinement steps to obtain a higher-quality scene representation. The resulting 3D asset can be immediately rendered from novel viewpoints, edited, transmitted, or stored as a compact scene-level memory. Compared with conventional image or video capture, this pipeline records not only appearance but also a renderable spatial representation of the scene.

We view this as an initial step toward a new imaging paradigm. Traditional cameras capture 2D observations, leaving geometry, view synthesis, and scene understanding to offline processing. An edge 3D reconstruction camera instead treats 3D representation as the native output of capture. ForeSplat contributes to this direction by making the predict-then-refine pipeline more compatible with compact models and limited compute budgets, thereby bringing high-quality feed-forward 3DGS reconstruction closer to practical, portable, and interactive deployment.

%% sec/6_conclusion.tex
\section{Conclusion and Limitations}
\label{section:conclusion}

ForeSplat addresses a central training-deployment mismatch in feed-forward 3DGS: practical systems are often used in a predict-then-refine regime, yet standard training optimizes only immediate rendering quality. By training the Gaussian prediction head with the optimization-aware \emph{MetaGrad} rule, ForeSplat encourages feed-forward models to produce initializations that are easier for a downstream optimizer to improve. Across multiple backbones, this shift consistently yields better post-optimization trajectories under matched budgets, even when the zero-step rendering quality is lower.

Our current study is limited to fixed-topology post-optimization, with densification, cloning, splitting, and pruning disabled. Extending optimization-aware training to refinement pipelines incorporating structural 3DGS operations, additional optimizer adaptation, and broader Gaussian settings remains important future work. More broadly, these results suggest that learning better initializations is a practical path toward bridging fast feed-forward prediction and high-quality iterative refinement.

{
    \small
    \bibliographystyle{ieeenat_fullname}
    \bibliography{main}
}

\newpage

\twocolumn[%
\begin{center}
{\Large \textbf{ForeSplat: Optimization-Aware Foresight for Feed-Forward 3D Gaussian Splatting}}

\vspace{0.5em}

{\Large Supplementary Material}
\end{center}

\vspace{1em}
]

\section{MetaGrad Pseudocode}
\label{suppl:pseudo}

\Cref{alg:metagrad} summarizes one iteration of the MetaGrad training rule within the ForeSplat framework on a single training tuple $\mathcal{I}$.
The notation matches Section 3.4.

\begin{algorithm}[!h]
\DontPrintSemicolon
\KwData{
    Tuple $\mathcal{I}$; 
    weights $\Theta$ of FF-3DGS network $f_\Theta$; 
    host loss $L_A$;
    max post opt step $K_{\max}$; 
    anchor stride $\Delta$;
    inner/outer LRs $\eta,\eta_{\mathrm{out}}$; 
    hyper parameter $\lambda\in[0,1]$;
    stop-grad $\mathrm{sg}(\cdot)$; 
    $\mathbf{Adam}(\cdot)$: one Adam step.
}

\KwResult{Updated weights $\Theta$.}
$G_0 \gets f_\Theta(\mathcal{I})$\;
$L_0 \gets L_A(G_0)$\;
$K_{\mathrm{rand}} \sim \mathcal{U}\{K_{\min},\dots,K_{\max}\}$\;
$G_{opt} \gets \mathrm{sg}(G_0)$\;
$Sum_{grad}\gets \mathbf{0}$, \quad 
$N \gets 0$\;
\For{$k \gets 1$ \KwTo $K_{\mathrm{rand}}$}{
    $G_{opt} \gets \mathbf{Adam}(G_{opt}, \nabla_G L_A(G), \eta)$\;
    \If{$k \bmod \Delta = 0$}{
        $Sum_{grad}\gets Sum_{grad} + \nabla_{G} L_A(\mathrm{sg}(G_{opt}))$\;
        $N \gets N + 1$\;
    }
}
$\hat{g}_0 \gets Sum_{grad} / N$\;
$g^{\mathrm{imm}} \gets \nabla_\Theta L_0$\;
$g^{\mathrm{meta}} \gets \hat{g}_0 \cdot \partial f_\Theta(\mathcal{I}) / \partial \Theta$\;
$g_\Theta \gets \lambda\, g^{\mathrm{imm}} + (1-\lambda)\, g^{\mathrm{meta}}$\;
$\Theta \gets \mathbf{Adam}(\Theta, g_\Theta, \eta_{\mathrm{out}})$\;
\caption{MetaGrad training rule within ForeSplat: one training iteration.}
\label{alg:metagrad}
\end{algorithm}

\section{Pi3X Gaussian Head: Architecture and Training Protocol}
\label{suppl:gshead}
This section details the construction and pre-training of the Gaussian
head attached to the Pi3X backbone, which turns Pi3X into the FF-3DGS
network $f_\Theta$ used throughout Section 3.4.

\paragraph{Architecture.}
The Gaussian head is a lightweight DPT-style~\cite{ranftl2021dpt}
decoder grafted onto the frozen Pi3X transformer.
It taps the hidden states of four intermediate decoder layers
(indices $\{8, 18, 26, 34\}$ out of $36$), each projected from $2{,}048$
to $256$ channels by a $1\times1$ convolution and rescaled to a
coarse-to-fine pyramid via transposed and strided convolutions.
A stack of four feature-fusion blocks with residual convolutional units
progressively merges the pyramid up to the input resolution, and a
parallel $7\times7{+}3\times3$ convolutional skip path injects the
input RGB so that the head predicts residuals on top of the observed
pixel colors.
Two final $3\times3$ and $1\times1$ convolutions emit an
$83$-channel attribute map per pixel: $1$ opacity logit, $3$ scale
components, a $4$-dimensional rotation quaternion, and $75$ spherical
harmonics coefficients up to degree $4$.
Opacity is obtained as $\sigma(\cdot)$ modulated by the Pi3X confidence;
scales are pose-free, computed as $0.001 \cdot \mathrm{softplus}(\cdot)$
and clamped at $0.3$ so that they do not depend on per-pixel depth;
rotations are $\ell_2$-normalized quaternions; the SH DC term is added
as a residual to the input RGB, while higher-order coefficients are
damped by a degree-dependent factor.
The head contains $14.6$\,M parameters, i.e.\ about $1.06\%$ of the
full Pi3X-with-head model ($1{,}374.7$\,M).
To stabilize training, the last $1\times1$ convolution is zero
initialized, and biases are set such that the initial output
corresponds to a near-identity Gaussian (opacity logit $0$ and
identity quaternion).

\paragraph{Training protocol.}
We train only the Gaussian head, keeping the $1{,}360.0$\,M parameters
of the Pi3X backbone frozen, so that gradients flow exclusively
through the $14.6$\,M head parameters.
Training is performed on $8\times$ NVIDIA H20 GPUs with PyTorch DDP and FP16 mixed precision via
\texttt{GradScaler}, for $30{,}000$ iterations
($30$ epochs of $1{,}000$ iterations each).
We use RealEstate10K and DL3DV as training datasets, each iteration draws a single scene with $N=8$ views at
resolution $224\times224$, and one view is uniformly sampled as the
rendering target while the remaining views provide context.
The per-GPU batch size is $1$ with no gradient accumulation, giving an
effective batch size of $2$.
We use AdamW ($\beta = (0.9, 0.999)$) with peak learning rate
$1\times10^{-4}$, weight decay $0.05$, and gradient clipping at $1.0$;
the schedule is a $2{,}000$-step linear warm-up followed by cosine
decay.
The objective is a photometric reconstruction loss combining
pixel-space $\ell_2$ and perceptual VGG-LPIPS~\cite{zhang2018lpips},
\begin{equation}
\label{eq:gshead_loss}
\mathcal{L}_\text{head}
= \mathcal{L}_\text{mse} + 0.05\,\mathcal{L}_\text{lpips},
\end{equation}
evaluated between the rendered target view and the corresponding
ground-truth RGB.
The LPIPS weight is intentionally small to prevent the perceptual term
from dominating early training, when the head outputs are still close
to their initialization.
The resulting checkpoint provides the FF-3DGS network $f_\Theta$ that
serves as the input to the MetaGrad training rule
in~\cref{alg:metagrad}.

\section{Distill Pi3X: Architecture and Training Protocol}
\label{suppl:distill}
This section details the construction of \emph{Distill Pi3X}, the lightweight backbone introduced in Section 4.1.

\paragraph{Architecture.}
Distill Pi3X is obtained by distilling Pi3X---which couples a DINOv2 Large encoder with a $36$-layer Transformer decoder---into a student that pairs a DINOv2 Base encoder with a $24$-layer decoder.
The resulting model contains $612.8$\,M parameters, corresponding to approximately $44.6\%$ of Pi3X ($1389.4$\,M) and $51.5\%$ of AnySplat ($1{,}190.7$\,M).
The student preserves the three geometry heads of Pi3X (point, camera, confidence) and additionally carries a lightweight Gaussian head, so that a single forward pass produces both Pi3X-compatible geometry and renderable 3D Gaussians.
To bridge the dimensionality gap between the Base encoder and the Pi3X decoder, we insert a projection module consisting of a linear layer followed by a LayerNorm.
The encoder, decoder, and geometry heads are initialized from Pi3X, while the projection module and the Gaussian head are randomly initialized.
All components are trained jointly, with Pi3X serving as the teacher.

\paragraph{Training protocol.}
We reuse the dataset mixture and sampling protocol of Section 4.1 without modification.
The distill model is trained on $6\times$ NVIDIA H20 GPUs for $200$ epochs of $800$ iterations each.
We use the AdamW optimizer ($\beta = (0.9, 0.95)$) together with a OneCycle cosine schedule.
Per-group learning rates are set to $1 \times 10^{-4}$ for the Gaussian head, $5 \times 10^{-5}$ for the decoder and the projection module, and $5 \times 10^{-6}$ for the ViT-B encoder, the point head, and the camera modules; this configuration is chosen to preserve the geometric accuracy inherited from the teacher.
The total loss is given by
\begin{equation}
\label{eq:distill_total_loss}
\begin{aligned}
\mathcal{L}_\text{total}
={}& \mathcal{L}_\text{point}
+ \mathcal{L}_\text{normal}
+ 0.1\,\mathcal{L}_\text{camera}
+ 0.2\,\mathcal{L}_\text{feat} \\
& + 0.2\,\mathcal{L}_\text{out}
+ \mathcal{L}_\text{mse}
+ 0.05\,\mathcal{L}_\text{lpips},
\end{aligned}
\end{equation}
where $\mathcal{L}_\text{point}$, $\mathcal{L}_\text{normal}$, and $\mathcal{L}_\text{camera}$ align the student with the Pi3X teacher on local points, surface normals, and camera pose, respectively, with the camera term following the Pi3-style formulation ($\alpha = 100$, $\delta = 0.1$).
$\mathcal{L}_\text{feat}$ and $\mathcal{L}_\text{out}$ enforce alignment between the student and teacher hidden states at intermediate decoder layers and at the final decoder output.
$\mathcal{L}_\text{mse}$ and $\mathcal{L}_\text{lpips}$ are evaluated between the rendered image and the ground-truth RGB.

\paragraph{Accuracy trade-off.}
\Cref{tab:dl3dv_comparison} reports camera-pose accuracy on the DL3DV test split.
Distill Pi3X attains lower scores than Pi3, Pi3X, and AnySplat across all three metrics, with the gap most visible on $\text{AUC@30}$.
This drop is commensurate with the substantial reduction in parameter count---roughly $45\%$ of Pi3X and $51\%$ of AnySplat---and constitutes an expected cost of knowledge distillation.
At the same time, Distill Pi3X serves in our experiments as a representative instance of a more compact, geometrically lighter feed-forward backbone: the directionally consistent MetaGrad gains observed across all three backbones (Section 4.2) thus span a range of feed-forward model capacities, from the high-capacity AnySplat to the compact Distill Pi3X.

\begin{table}[t]
\centering
\caption{Camera pose estimation comparison on DL3DV test set. AUC@30, RRA@30, and RTA@30 measured against ground-truth poses.}
\setlength{\tabcolsep}{2pt}
\label{tab:dl3dv_comparison}
\begin{tabular}{lcccc}
\toprule
Method & \#Params & AUC@30$\uparrow$ & RRA@30$\uparrow$ & RTA@30$\uparrow$ \\
\midrule
Pi3             & 958.7M  & 0.859 & 0.971 & 0.955 \\
Pi3X            & 1389.4M & \textbf{0.880} & 0.982 & 0.957 \\
AnySplat        & 1190.7M & 0.879 & \textbf{0.985} & \textbf{0.988} \\
Distill Pi3X  & 612.8M  & 0.723 & 0.944 & 0.924 \\
\bottomrule
\end{tabular}
\end{table}

\section{Continuous Post-Optimization Trajectories}
\label{suppl:dynamics}

This section complements Sections~4.2 and~4.3 by reporting the
underlying post-optimization trajectories at a finer step resolution.
The figures visualize the evolution of the metrics over the same
$2{,}000$-step window summarized in Section~4.2, and provide the
complete $\lambda$ sweep trajectories on all three backbones.

\begin{figure*}[t]
    \centering
    \begin{subfigure}{\linewidth}
        \centering
        \includegraphics[width=0.9\linewidth]{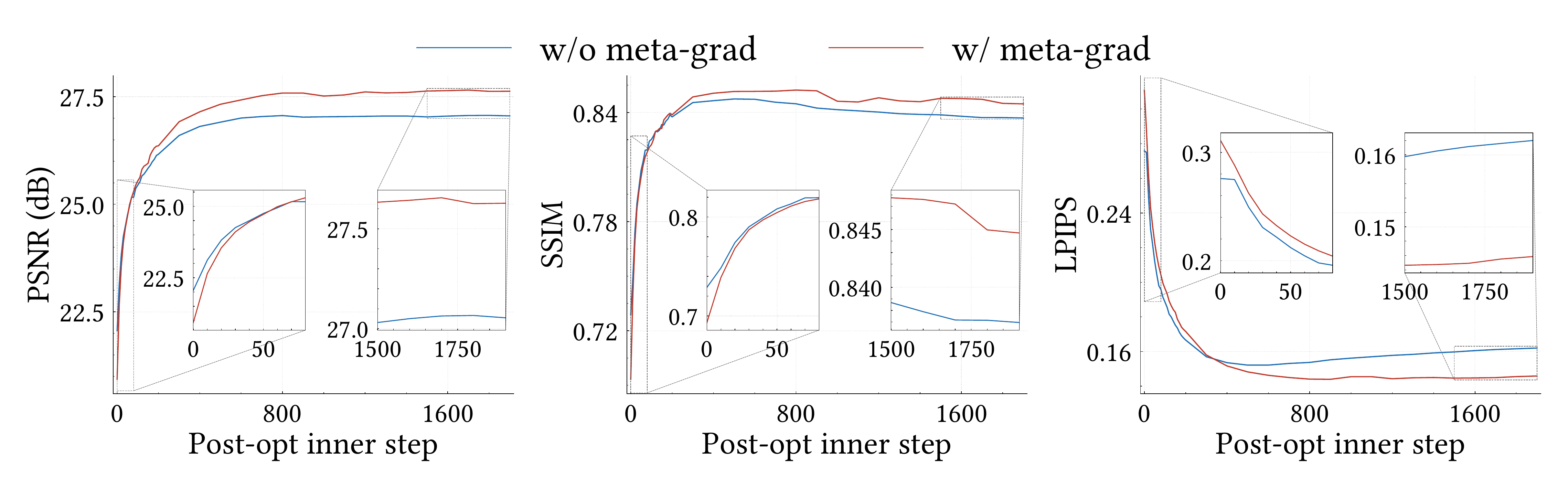}
        \caption{AnySplat.}
        \label{fig:main_anysplat}
    \end{subfigure}

    \vspace{0.4em}

    \begin{subfigure}{\linewidth}
        \centering
        \includegraphics[width=0.9\linewidth]{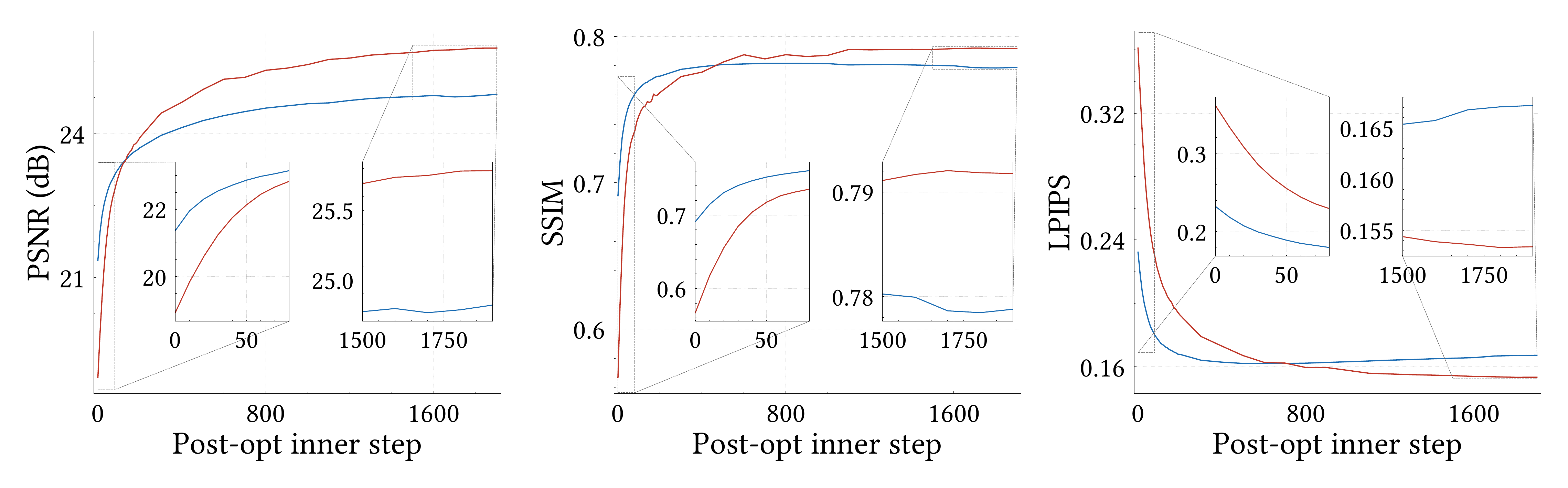}
        \caption{Pi3X.}
        \label{fig:main_pi3x}
    \end{subfigure}

    \vspace{0.4em}

    \begin{subfigure}{\linewidth}
        \centering
        \includegraphics[width=0.9\linewidth]{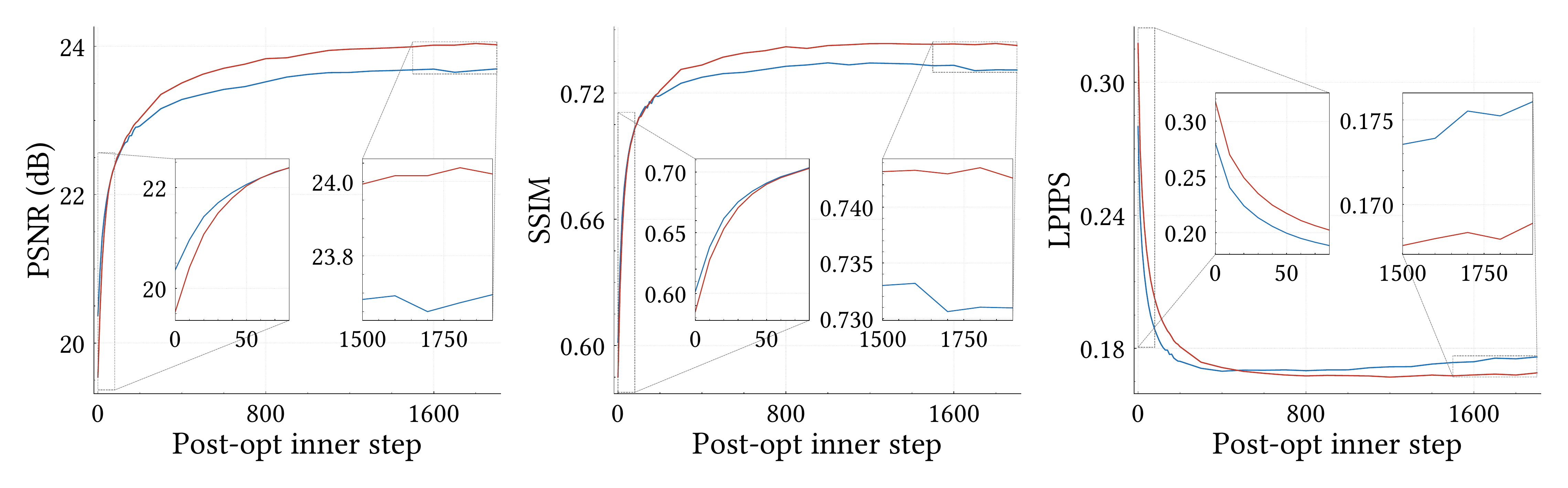}
        \caption{Distill Pi3X.}
        \label{fig:main_distill_pi3x}
    \end{subfigure}

    \caption{
    Post-optimization trajectories of the three backbones.
    (a)~AnySplat, (b)~Pi3X, and (c)~Distill Pi3X. Within each row, we report PSNR, SSIM, and LPIPS on held-out test views over the first $2{,}000$ post-optimization steps.
    }
    \label{fig:main_results}
    % \Description{fig: step - (psnr,ssim,lpips)}
\end{figure*}

\paragraph{Full trajectories under MetaGrad and vanilla.}
\Cref{fig:main_results} plots the PSNR, SSIM, and LPIPS curves of
AnySplat, Pi3X, and Distill Pi3X on held-out test views over the
first $2{,}000$ post-optimization steps. The trajectories evolve
smoothly between the steps reported in Section~4.2: the
MetaGrad-trained model lies slightly below the baseline in the early
phase of post-optimization, subsequently overtakes it, and remains
above the baseline throughout the rest of the interval. This
late-stage separation is sustained throughout the
$1{,}500$--$2{,}000$ step zoom-in insets, and the overall shape of
the curves remains qualitatively consistent across the three
backbones.

\begin{figure*}[t]
    \centering
    \begin{subfigure}{\linewidth}
        \centering
        \includegraphics[width=0.9\linewidth]{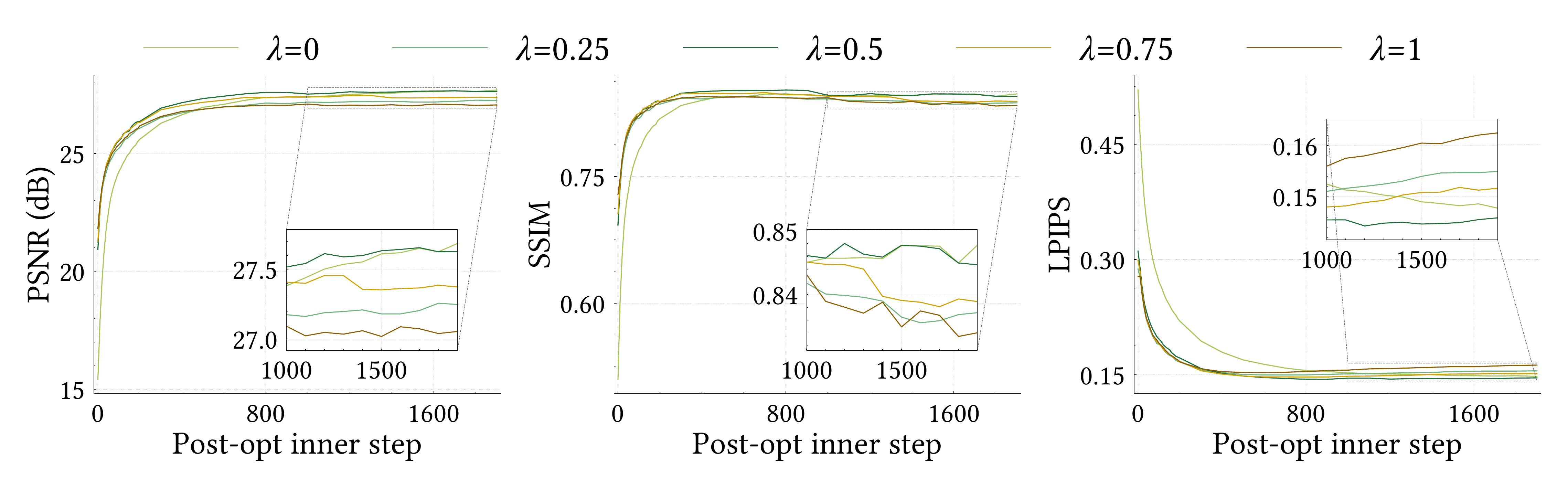}
        \caption{AnySplat.}
        \label{fig:lambda_anysplat}
    \end{subfigure}

    \vspace{0.4em}

    \begin{subfigure}{\linewidth}
        \centering
        \includegraphics[width=0.9\linewidth]{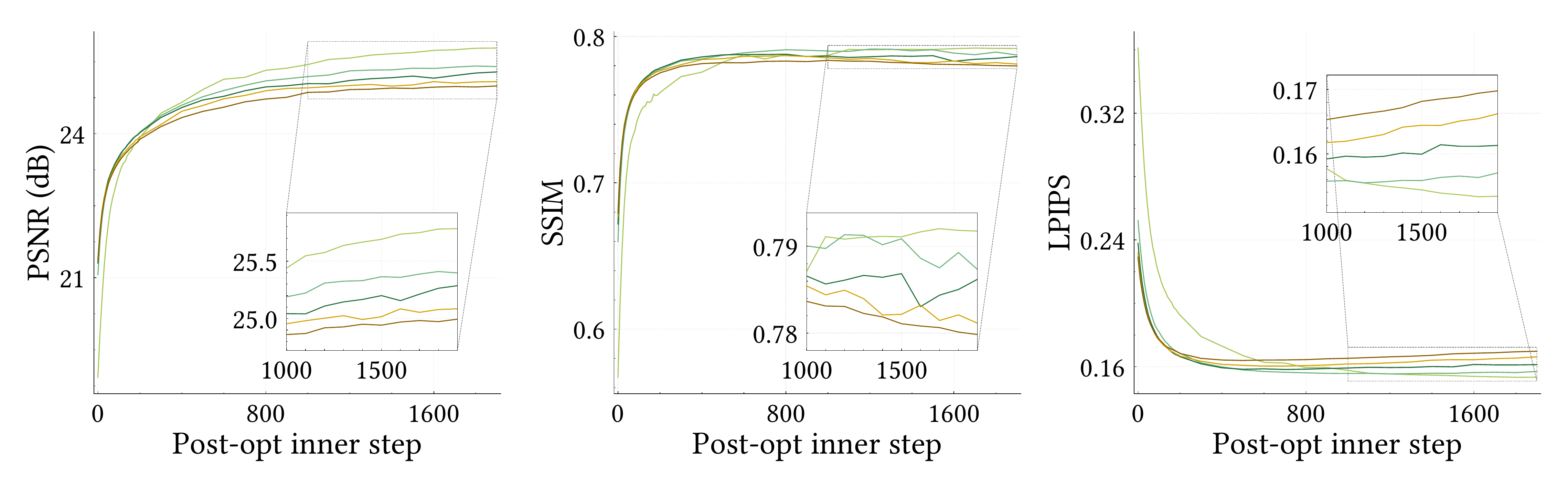}
        \caption{Pi3X.}
        \label{fig:lambda_pi3x}
    \end{subfigure}

    \vspace{0.4em}

    \begin{subfigure}{\linewidth}
        \centering
        \includegraphics[width=0.9\linewidth]{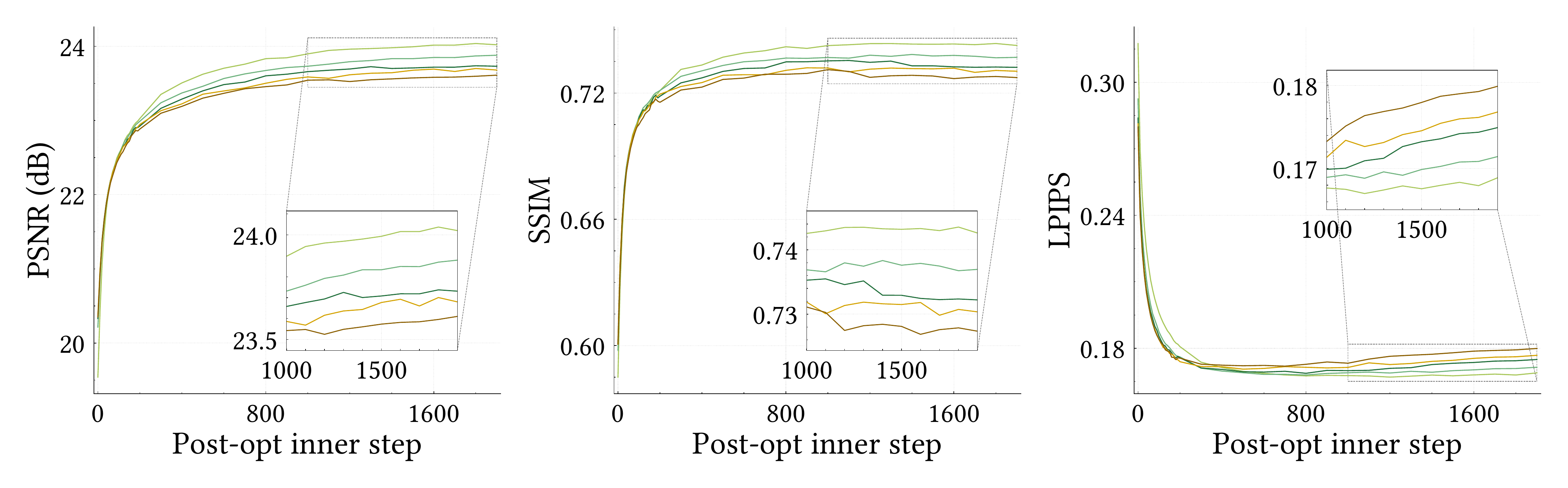}
        \caption{Distill Pi3X.}
        \label{fig:lambda_distill_pi3x}
    \end{subfigure}

    \caption{Effect of $\lambda$ on the three backbones.
    (a)~AnySplat, (b)~Pi3X, and (c)~Distill Pi3X. Within each row, we
    report PSNR, SSIM, and LPIPS measured at $2{,}000$
    post-optimization steps as a function of the loss-balancing
    coefficient $\lambda \in \{0.0, 0.25, 0.5, 0.75, 1.0\}$. The two
    endpoints correspond to vanilla supervised fine-tuning
    ($\lambda = 1.0$) and a pure-meta variant ($\lambda = 0.0$),
    respectively. 
    Across all three backbones, the best-performing
    region lies away from the purely supervised endpoint and remains broadly similar across backbones.}
    \label{fig:lambda_sweep}
    % \Description{fig: ablation sweep lambda}
\end{figure*}

\paragraph{$\lambda$ sweep on all three backbones.}
The hyperparameter $\lambda$ controls the relative weighting between
the gradient of the original supervised loss and the meta-gradient
in the training objective, with a smaller $\lambda$ assigning a
larger share to the meta-gradient in the resulting combined
gradient. \Cref{fig:lambda_sweep} reports the complete
post-optimization trajectories of all three backbones under
$\lambda \in \{0.0, 0.25, 0.5, 0.75, 1.0\}$. On each backbone, all
MetaGrad variants exhibit the same qualitative trajectory shape as
above---initially below the baseline and subsequently above
it---while the value of $\lambda$ exhibits a stable correspondence
with the resulting reconstruction quality: the PSNR, SSIM, and LPIPS
curves largely retain the same relative ordering throughout the
post-optimization horizon, and this ordering is broadly consistent
across the three backbones.

\end{document}